\def\SCULPTArxiv{1}
\documentclass{article}
\ifx\pdfminorversion\undefined\else\pdfminorversion=7\fi

\PassOptionsToPackage{table}{xcolor}
\usepackage{iclr2027_conference,times}
\usepackage{float}
\usepackage{subcaption}
\usepackage[utf8]{inputenc}
\usepackage[T1]{fontenc}
\usepackage{url}
\usepackage{graphicx}
\usepackage{wrapfig}
\usepackage{placeins}
\usepackage{flafter}
\usepackage{booktabs}
\usepackage{multirow}
\usepackage{tabularx}
\usepackage{dashrule}
\usepackage{amsmath}
\usepackage{amsfonts}
\usepackage{nicefrac}
\usepackage{microtype}
\usepackage{xcolor}
\usepackage[most]{tcolorbox}
\usepackage{etoolbox}
\usepackage{colortbl}
\usepackage[breaklinks=true]{hyperref}

\definecolor{easybg}{RGB}{219,235,218}
\definecolor{mediumbg}{RGB}{239,231,193}
\definecolor{hardbg}{RGB}{240,220,199}
\definecolor{toponebg}{RGB}{169,208,142}
\definecolor{toptwobg}{RGB}{190,219,160}
\definecolor{topthreebg}{RGB}{209,229,176}
\definecolor{topfourbg}{RGB}{225,236,186}
\definecolor{topfivebg}{RGB}{239,231,193}
\newcommand{\topone}[1]{\cellcolor{toponebg}\textbf{#1}}
\newcommand{\toptwo}[1]{\cellcolor{toptwobg}#1}
\newcommand{\topthree}[1]{\cellcolor{topthreebg}#1}
\newcommand{\topfour}[1]{\cellcolor{topfourbg}#1}
\newcommand{\topfive}[1]{\cellcolor{topfivebg}#1}
\newcommand{\tok}[1]{\texttt{\textless\detokenize{#1}\textgreater}}
\newcommand{\slotrep}[2]{\tok{#1}$\times #2$}

\definecolor{stageblue}{RGB}{33,72,170}
\definecolor{notegreen}{RGB}{20,120,60}
\definecolor{notegray}{RGB}{90,90,90}
\definecolor{konebg}{RGB}{244,239,255}
\definecolor{koneframe}{RGB}{121,94,214}
\definecolor{ktwobg}{RGB}{236,244,255}
\definecolor{ktwoframe}{RGB}{70,120,210}
\definecolor{kthreebg}{RGB}{236,249,244}
\definecolor{kthreeframe}{RGB}{55,150,110}
\definecolor{kfourbg}{RGB}{255,245,232}
\definecolor{kfourframe}{RGB}{220,140,40}
\definecolor{kfivebg}{RGB}{255,239,245}
\definecolor{kfiveframe}{RGB}{205,95,145}

\definecolor{impbg}{RGB}{245,243,255}
\definecolor{impframe}{RGB}{90,80,220}
\title{SCULPT-VLA: Learning Structured Control through Staged Action Grounding}

\definecolor{trainbg}{RGB}{232,245,233}
\definecolor{trainfg}{RGB}{46,125,50}
\definecolor{freezebg}{RGB}{252,235,235}
\definecolor{freezefg}{RGB}{198,40,40}
\definecolor{lowlrbf}{RGB}{235,242,255}
\definecolor{lowlrfg}{RGB}{49,91,182}

\newcommand{\trainable}{\cellcolor{trainbg}\textcolor{trainfg}{$\checkmark$}}
\newcommand{\frozenm}{\cellcolor{freezebg}\textcolor{freezefg}{$\times$}}
\newcommand{\lowlr}{\cellcolor{lowlrbf}\textcolor{lowlrfg}{$\triangle$}}

\hypersetup{pdftitle={SCULPT-VLA: Learning Structured Control through Staged Action Grounding}}
\ifdefined\SCULPTArxiv
  \iclrfinalcopy
  \author{%
Wenbo Li$^{1}$ \quad Yiteng Chen$^{1}$ \quad Wei Zhang$^{1}$\\
\bfseries Wenhao Li$^{1}$ \quad Jun Yang$^{2}$ \quad Qingyao Wu$^{1,*}$\\[0.5em]
\normalfont $^{1}$School of Software Engineering, South China University of Technology\\
\normalfont Guangzhou, China\\
\normalfont $^{2}$Yuanwu Technology, Shenzhen, China\\[0.3em]
\normalfont $^{*}$Corresponding author: \texttt{qyw@scut.edu.cn}
}
\hypersetup{pdfauthor={Wenbo Li, Yiteng Chen, Wei Zhang, Wenhao Li, Jun Yang, Qingyao Wu}}

\else
  \author{Anonymous authors}
  \hypersetup{pdfauthor={Anonymous authors}}
\fi
\begin{document}
\maketitle
\ifdefined\SCULPTArxiv
  \lhead{SCULPT-VLA}
\fi
\begin{abstract}
Vision-language-action (VLA) policies increasingly incorporate structured
intermediate supervision beyond action labels.
Yet specifying what an intermediate representation should encode leaves
open how action prediction learns to depend on it.
We introduce \textbf{SCULPT-VLA}, a policy that learns structured
control through staged action grounding.
Its action-conditioning state comprises complementary factors for
task progression, scene dynamics, and spatial grounding.
Training first forms these factors with teacher scaffolds, then
grounds coarse action prediction through their composition as scaffold
inputs are withdrawn. Direct perceptual access is subsequently
restored for continuous refinement, combining the learned state with
perceptual detail. The curriculum separates learning to condition actions
on structure from refining continuous control.
Deployment requires neither teachers nor discrete-action autoregression.
SCULPT-VLA achieves higher average success than shared-backbone baselines
on LIBERO, SimplerEnv-WidowX, and RoboTwin 2.0 Full.
On SimplerEnv-WidowX, final success is 83.5\%, versus 71.3\% when
Stage-II action learning directly accesses vision and language.
Across four physical robot tasks, average success under the tested
distribution shifts reaches 58.1\%, compared with 45.6\% for $\pi_{0.5}$.
Training ablations and factor-wise interventions support the staged
design and show that the learned state continues to contribute to
control after direct perceptual access is restored.
\end{abstract}

\section{Introduction}
As a robot stacks a cup, the relevant control problem changes from
transporting the object to aligning its rim and releasing the gripper.
The objects and instruction may stay the same, yet the next action
depends on a changing combination of task progress, relative geometry,
and interaction-induced motion. Vision-language-action (VLA) models
bring broad semantic knowledge to these decisions
\citep{brohan2023rt2,kim2024openvla,black2024pi0}.
The opportunity is to turn this knowledge into a control state that
tracks the changing demands of manipulation.

Recent policies enrich action learning with intermediate supervision
beyond demonstrated actions.
Task decompositions and affordance plans organize execution and
interaction \citep{physicalintelligence2025pi05,zawalski2024robotic,nasiriany2024rtaffordance},
while spatial and predictive representations supply geometric and
transition information \citep{qu2025spatialvla,bu2025univla,qiu2026agra}.
World-action models and structured feature objectives further connect
these signals to action generation \citep{bai2026bridgewa,zheng2026gift}.
These advances raise a complementary question about policy learning
(Fig.~\ref{fig:main_arch}): \emph{how should action learning establish
a functional role for this structure?}

The conditioning pathways that support policy learning may differ from
those needed at deployment.
When the action decoder has unrestricted access to multimodal context
from the outset, joint optimization must learn both the representations
and how to condition actions on them.
Conditioning actions exclusively on an intermediate state instead
requires that state to preserve every detail needed for continuous control.
This motivates a staged approach: first ground action learning through
a structured state, then restore direct perceptual access for refinement.
The central question is whether the resulting state continues to
contribute when the final policy can also use perceptual detail.

\textbf{SCULPT-VLA} realizes this principle through
\emph{staged action grounding}, which shapes the dependence of action
prediction on the structured control state during training.
A shared backbone organizes this state into task-progression,
scene-dynamics, and spatial-grounding factors, distinct from the
action readout.
Teacher scaffolds first establish factor roles. Coarse action
learning then draws its non-action context from their composition
as scaffold inputs are withdrawn.
Continuous refinement then restores direct perceptual access,
conditioning on factor states, action readout, and visual detail.
The deployed policy forms its control state from observations and
instructions, without teacher encoders or discrete-action autoregression.

Evaluations on LIBERO, SimplerEnv-WidowX, and RoboTwin 2.0 Full show
higher average success than shared-backbone baselines.
Four physical robot tasks further demonstrate gains under the tested
distribution shifts. Training controls examine how this capability is
acquired, and final-state interventions
test whether the factors still contribute after direct access is
restored. We contribute
(1) a \textbf{structured control state} that organizes task progression,
scene dynamics, and spatial grounding separately from action readout;
(2) \textbf{staged action grounding}, a training method linking
factor formation to continuous control through staged changes in
information access; and
(3) \textbf{simulation, physical-robot, and functional-intervention
evidence} supporting this design and the learned state's contribution
to deployed control.

\section{Related Work}
\subsection{Vision-Language-Action Models}
VLA models transfer pretrained vision-language representations to
robot control through discrete action prediction
\citep{brohan2023rt2,kim2024openvla,pertsch2025fast} or continuous
action generation \citep{black2024pi0}.
Their action interfaces increasingly structure multimodal representations
before decoding: bridge representations and lightweight adapters
connect pretrained backbones to action heads
\citep{li2025bridgevla,yang2025instructvla,wang2025vlaadapter}, while
Action QFormer constructs an instruction-conditioned, query-based
representation for action decoding \citep{ji2026actionqformer}.

\subsection{World Models and Predictive Robot Learning}
Predictive policies learn from anticipated observations and scene
changes \citep{dreamvla25,zhu2026deltavla}, while latent-action models
extract compact transition representations
\citep{bu2025univla,cai2026ssmvla}.
Recent world-action models explicitly address how predictive features
support action generation. AGRA aligns video-model features with
spatially coherent semantic representations to improve the
world-action interface \citep{qiu2026agra}.
Bridge-WA distills future, change, and motion-flow priors into a
policy, conditions action generation on these priors, and removes
the teacher at inference \citep{bai2026bridgewa}.
These methods establish predictive representations as a source of
control-relevant information beyond current observations.

\subsection{Structured Intermediate Representations for Robot Control}
Affordance plans, geometric representations, and structured feature
supervision organize control-relevant information
\citep{nasiriany2024rtaffordance,qu2025spatialvla,zheng2026gift}.
Reasoning policies connect task semantics to actions through explicit
subtasks or reasoning traces
\citep{zawalski2024robotic,physicalintelligence2025pi05,zhao2025cotvla,wang2026vlathinker}
and latent reasoning states
\citep{bai2026laravla,liu2026last0,huang2026fastthinkact}.
Latent Semantic Scaffolding aligns action-token representations
with phase-specific reasoning targets during training and discards
the alignment head at inference \citep{li2026lss}.
Counterfactual VLA analyses also examine whether encoded instruction
information affects behavior \citep{grant2026not,fang2026vision}.
SCULPT-VLA examines how structured representations acquire a
functional role during action learning.
Factor formation is followed by action grounding through the composed
state, then refinement with direct perceptual access restored.

\section{Method}
\begin{figure}[!htbp]
  \centering
  \includegraphics[width=\linewidth]{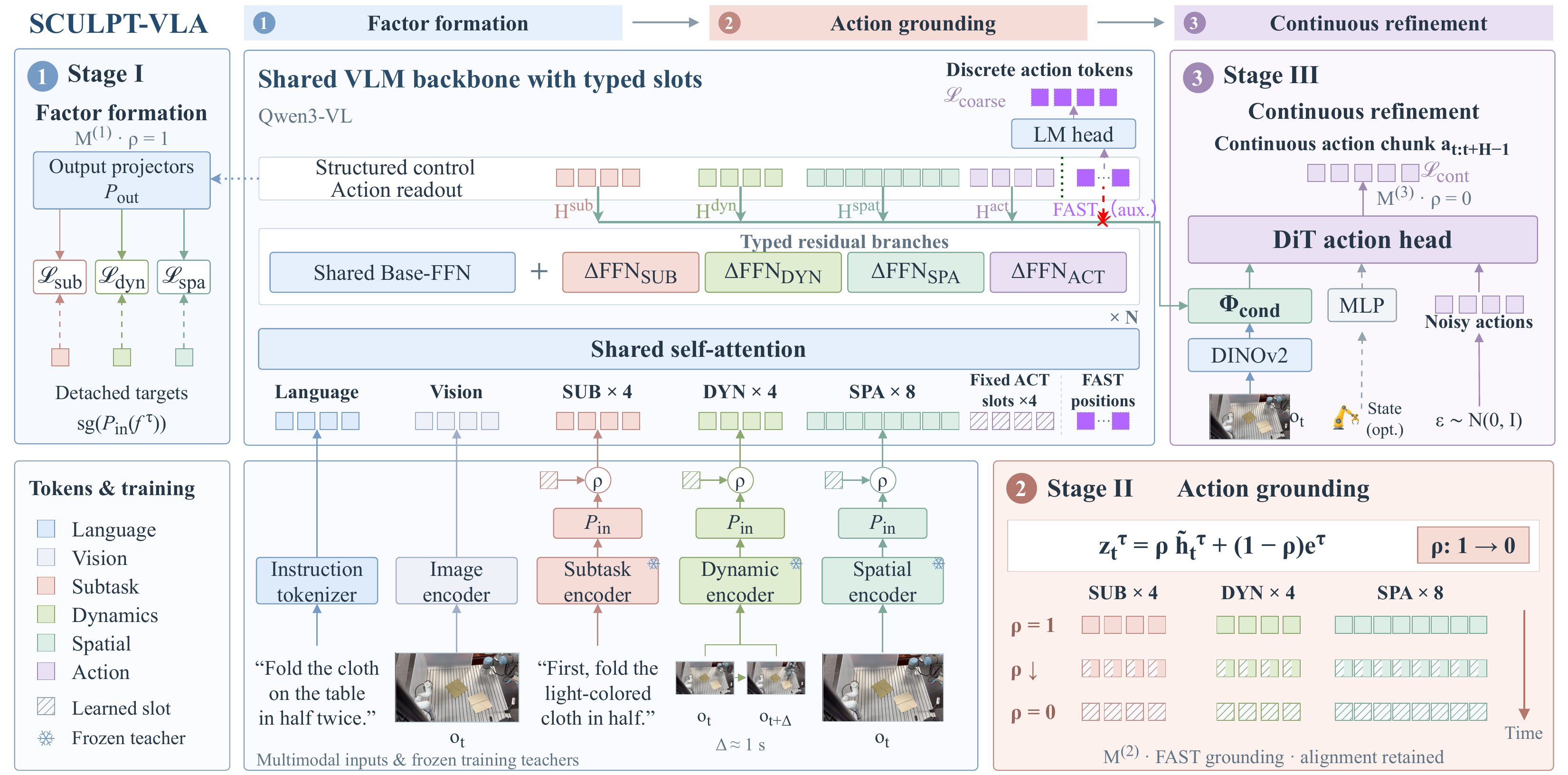}
  \caption{SCULPT-VLA overview. Stage~I forms teacher-aligned factors;
  Stage~II withdraws scaffold inputs while grounding FAST prediction
  through interacting factors; Stage~III restores direct perceptual
  access for continuous refinement. The continuous head jointly uses
  factor states, fixed action readout, and DINOv2 features.}
  \label{fig:main_arch}
\end{figure}

SCULPT-VLA learns structured control through staged action grounding
(Fig.~\ref{fig:main_arch}). Structured control denotes action
prediction conditioned on complementary control factors explicitly
trained for use in action learning.
The method separates factor formation from action readout and schedules
their interaction during training.

\subsection{Structured Control State}
\label{sec:preliminaries}
\paragraph{Inputs and control factors.}
At time $t$, the policy receives an observation $o_t$ and instruction
$x$, denoted by $\mathbf q_t=(o_t,x)$, and predicts an action chunk
$\mathbf a_{t:t+H-1}\in\mathbb R^{H\times d_a}$,
where $H$ is the action horizon and $d_a$ the action dimension.
We structure the action-conditioning state around three
control-oriented factors: \textbf{task progression} ($\mathrm{sub}$),
\textbf{scene dynamics} ($\mathrm{dyn}$), and \textbf{spatial grounding}
($\mathrm{spa}$). They address recurring requirements of manipulation:
tracking the active subtask, accounting for interaction-induced
changes, and localizing relevant geometry.

\paragraph{Structured factors and action readout.}
We implement the factors as typed latent slots, fixed token positions
whose contextualized states form the structured control state.
We denote these states by
$\mathbf h_t^{\mathrm{sub}},\mathbf h_t^{\mathrm{dyn}},
\mathbf h_t^{\mathrm{spa}}$.
Additional fixed action-side slots $\mathbf h_t^{\mathrm{act}}$
serve as \emph{action readout tokens}, aggregating context for
action decoding. These differ from the FAST positions
$\mathbf Z_t^{\mathrm{fast}}$ used to predict a discrete action
sequence \citep{pertsch2025fast}.
For discrete prediction, the action group uses causal attention at every
layer, with fixed readout preceding FAST positions and unable to access
the teacher-forced FAST suffix.
We denote the complete input sequence by $\mathbf Z_t^{(0)}$.
Group definitions, slot counts, and token layouts are given in
Appendix~\ref{sec:appendix_training_details}.

\paragraph{Teacher-guided factor formation.}
Complementary frozen teachers provide training-time scaffolds:
Qwen3-Embedding supplies instruction-aware task semantics,
UniVLA latent actions provide inverse-dynamics cues, and
Depth Anything V2 supplies current-scene geometry
\citep{zhang2025qwen3,bu2025univla,yang2024depth}.
For $\tau\in\{\mathrm{sub},\mathrm{dyn},\mathrm{spa}\}$,
the feature $\mathbf f_t^\tau$ is mapped to a scaffold input
$\tilde{\mathbf h}_t^\tau=P_{\mathrm{in}}^\tau
(\mathrm{sg}(\mathbf f_t^\tau))$.
The student representation is aligned to the detached reference
$\bar{\mathbf h}_t^\tau=\mathrm{sg}(\tilde{\mathbf h}_t^\tau)$:
\begin{equation}
\mathcal L_\tau=\frac{1}{|\mathbf h_t^\tau|}\sum_k
\left\|P_{\mathrm{out}}^\tau(\mathbf h_{t,k}^\tau)
-\bar{\mathbf h}_{t,k}^\tau\right\|_2^2
\label{eq:slot_align}
\end{equation}
Here $\mathrm{sg}$ stops gradients, and both projectors are trainable
while teacher encoders remain frozen. The scaffolds establish
factor-specific targets; the following curriculum trains those
factors to support action without teacher inputs.

\label{sec:architecture}
\paragraph{Factor-specific computation.}
Self-attention and the pretrained FFN remain shared.
Small type-specific residual FFNs preserve a dedicated computational
path for each factor and the action readout. For token $i$ at layer $l$,
\begin{equation}
\widetilde{\mathbf z}_l^{(i)}=\mathbf z_l^{(i)}
+\mathrm{FFN}_{\mathrm{base}}^{(l)}(\mathbf z_l^{(i)})
+\mathbb I[\tau(i)\ne\emptyset]\,
\Delta\mathrm{FFN}_{\tau(i)}^{(l)}(\mathbf z_l^{(i)})
\label{eq:typed_ffn}
\end{equation}
Vision and language tokens have $\tau(i)=\emptyset$ and retain the
base path. Token types select their residual FFNs deterministically.

\subsection{Staged Action Grounding}
\label{sec:training}
\begin{figure}[!htbp]
  \centering
  \includegraphics[width=\linewidth]{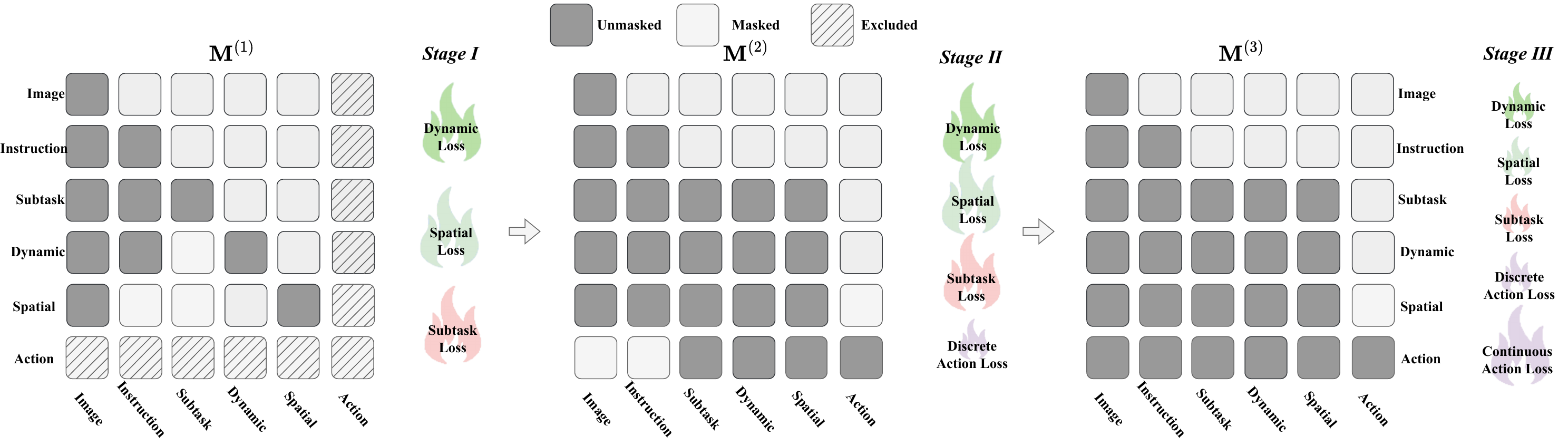}
  \caption{Stage-specific attention masks and objectives.
  Rows are query groups; columns are key groups. Dark gray permits
  attention, light gray blocks it, and hatching marks excluded groups.
  The discrete Action group uses causal attention over fixed readout
  followed by FAST positions. The continuous Action group contains only fixed readout.}
  \label{fig:mask}
\end{figure}

Action grounding trains the structured state to support action
prediction. The three stages regulate factor interaction and
action conditioning through the attention patterns in Fig.~\ref{fig:mask}.

\paragraph{Stage I: Forming structured control states.}
Under $\mathbf M^{(1)}$, task and dynamics factors attend to vision,
language, and their own group; the spatial factor attends to vision
and its own group, without language access.
Cross-factor attention and the action group are inactive, allowing
factor-specific alignment to teacher targets.

\paragraph{Stage II: Grounding actions through structured states.}
Under $\mathbf M^{(2)}$, the factors interact bidirectionally, while
action readout and FAST positions can read these factors but cannot
directly read vision or language. The composed factor state therefore
provides the non-action context for FAST prediction.

\label{para:scaffold_masking}
During Stage~II, teacher-derived inputs are replaced gradually by
learned query embeddings $\mathbf e_k^\tau$:
\begin{equation}
\mathbf z_{t,k}^{\tau,(0)}=\rho(r)\tilde{\mathbf h}_{t,k}^\tau
+(1-\rho(r))\mathbf e_k^\tau,
\label{eq:scaffold_input_mask}
\end{equation}
\begin{equation}
\rho(r)=\operatorname{clip}\left(
1-\frac{r-r_{\mathrm{anneal}}^{\mathrm{start}}}
{r_{\mathrm{anneal}}^{\mathrm{end}}-r_{\mathrm{anneal}}^{\mathrm{start}}},
0,1\right)
\label{eq:anneal}
\end{equation}
Here $r$ is the training step. Alignment remains active as
$\rho$ decreases to zero. By the end of annealing, the backbone infers
factor states from observations and instructions.

\paragraph{Stage III: Refining continuous control with restored perception.}
Under $\mathbf M^{(3)}$, action-side positions regain direct perceptual
and language access for continuous refinement; factor-group visibility
remains unchanged.
The continuous-control pass contains factor and fixed readout positions,
with no teacher-forced FAST suffix.
Weak FAST supervision uses a separate autoregressive auxiliary path.

\subsection{Training Objectives and Deployment}
\paragraph{Structured state formation.}
The Stage~I objective combines the factor-alignment losses:
\begin{equation}
\mathcal L^{(1)}=\sum_{\tau\in\{\mathrm{sub},\mathrm{dyn},\mathrm{spa}\}}
\lambda_\tau^{(1)}\mathcal L_\tau
\label{eq:stage1_obj}
\end{equation}

\paragraph{Coarse action grounding.}
FAST provides coarse action supervision in Stage~II and a weak auxiliary
objective through a separate discrete pass in Stage~III.
We tokenize the demonstrated action chunk into a sequence
$\mathbf c_t$ and train its autoregressive prediction:
\begin{equation}
\mathcal L_{\mathrm{coarse}}^{(s)}=
\mathcal L_{\mathrm{FAST}}(\mathbf c_t\mid\mathbf Z_t^{(0)},\mathbf M^{(s)}),
\qquad s\in\{2,3\}
\label{eq:coarse_action_obj}
\end{equation}
Stage~II combines this loss with continued factor alignment:
\begin{equation}
\mathcal L^{(2)}=\sum_\tau\lambda_\tau^{(2)}(r)\mathcal L_\tau
+\lambda_{\mathrm{coarse}}^{(2)}(r)\mathcal L_{\mathrm{coarse}}^{(2)}
\label{eq:stage2_obj}
\end{equation}

\paragraph{Continuous refinement.}
A flow-matching head \citep{black2024pi0} conditions on the final
factor representations, action readout states, and a DINOv2
visual-detail branch \citep{oquab2023dinov2}:
\begin{equation}
\mathbf u_t=\Phi_{\mathrm{cond}}(
\mathbf H_t^{\mathrm{slot},(L)},\mathbf H_t^{\mathrm{act},(L)},
\mathbf E_t^{\mathrm{DINOv2}})
\label{eq:action_condition}
\end{equation}
Here $\mathbf H_t^{\mathrm{slot},(L)}$ denotes final-layer structured factor
states, $\mathbf H_t^{\mathrm{act},(L)}$ the fixed action-readout states,
and $\mathbf E_t^{\mathrm{DINOv2}}$ the visual-detail features.
For Gaussian noise $\boldsymbol\epsilon$ and
$\alpha\sim\mathcal U(0,1)$, let
$\mathbf a_\alpha=(1-\alpha)\boldsymbol\epsilon+\alpha\mathbf a_{t:t+H-1}$.
The action expert $v_\theta$ learns the corresponding velocity field:
\begin{equation}
\mathcal L_{\mathrm{cont}}=\mathbb E_{\mathbf a,\boldsymbol\epsilon,\alpha}
\left[\left\|v_\theta(\mathbf a_\alpha,\alpha\mid\mathbf u_t)
-(\mathbf a_{t:t+H-1}-\boldsymbol\epsilon)\right\|_2^2\right]
\label{eq:cont_action_obj}
\end{equation}
Stage~III combines flow matching with weak alignment and the
separate FAST auxiliary loss:
\begin{equation}
\mathcal L^{(3)}=\lambda_{\mathrm{cont}}^{(3)}\mathcal L_{\mathrm{cont}}
+\sum_\tau\eta_\tau\mathcal L_\tau
+\eta_{\mathrm{coarse}}\mathcal L_{\mathrm{coarse}}^{(3)},
\qquad \eta_\cdot\ll1
\label{eq:stage3_obj}
\end{equation}
\paragraph{Deployment.}
At inference, learned query embeddings initialize the factor and
action readout positions, and the
backbone forms the control state without teacher encoders or alignment
projectors. The continuous head decodes action chunks from these states
and DINOv2 visual-detail features, without FAST autoregression.
Implementation and training settings are in
Appendix~\ref{sec:appendix_training_details}.

\section{Experiments}
We evaluated whether SCULPT-VLA converts structured supervision into
closed-loop visuomotor control across simulation and physical manipulation.
Training controls tested the learning pathway, and factor-wise
interventions examined the structured state's role after direct
perceptual access was restored.

\subsection{Evaluation Setup}
\label{sec:exp-setup}
We evaluated on the four LIBERO suites \citep{liu2023libero},
SimplerEnv-WidowX \citep{li2024evaluating}, RoboTwin 2.0 Full
\citep{chen2025robotwin}, and four real-world tabletop tasks.
For LIBERO and SimplerEnv, training uses LIBERO-SUBTASK and
BRIDGE-SUBTASK (Appendix~\ref{sec:video2tasks_pipeline}).
Our main results on these benchmarks use three seeds and 50 simulation
trials per task and seed.
Four Qwen3-VL baselines share the backbone, action data, action horizon,
and StarVLA evaluation pipeline \citep{community2026starvla}, with
FAST, OFT, $\pi_0$-style flow, or GR00T-style action interfaces.
Published results from other training stacks provide additional context
and are listed separately.
We report closed-loop success and mean open-loop score (mOLS), the
average of four thresholded action-error scores, with higher values
indicating better agreement with demonstration actions
(Appendix~\ref{app:open-loop}).
Table~\ref{tab:libero_robotwin} reports RoboTwin 2.0 Full results:
joint training over 50 tasks with 50 clean and 500 randomized
demonstrations per task, followed by Clean and Random evaluation.
Evaluation protocols appear in
Appendices~\ref{app:baseline_parity} and~\ref{app:extended_results}.

\subsection{Main Results}
\label{sec:exp-main}
\paragraph{Simulation.}
SCULPT-VLA achieves 98.5\% average LIBERO success and 96.9\% on the
Long suite, compared with 96.6\% and 93.8\% for Qwen3-VL-OFT, the
strongest shared-backbone baseline on LIBERO (Table~\ref{tab:main_results}).
On SimplerEnv-WidowX, it reaches 83.5\%, exceeding Qwen3-VL-GR00T
at 65.3\% and the direct continuous Qwen3-VL-PI policy at 60.9\%.
The gain over Qwen3-VL-GR00T is largest on Stack, where success rises
from 18.8\% to 75.3\%.
On RoboTwin 2.0 Full, SCULPT-VLA reaches 92.2\% average success,
exceeding Qwen3-VL-PI, the strongest shared-backbone StarVLA action-head
variant in Table~\ref{tab:libero_robotwin}.
Since LIBERO scores are close to saturation, the following ablations
focus on SimplerEnv.
% Main Table 1: LIBERO and SimplerEnv-WidowX.
% Numeric provenance: doc/TABLE1_SOURCES.md; pre-split values: audit/table_split.
\begin{table}[!htbp]
\centering
\caption{Main results on LIBERO and SimplerEnv-WidowX (success, \%).
Published methods retain their reported protocols. Shared-backbone variants
use Qwen3-VL in StarVLA~\citep{community2026starvla}.
SCULPT-VLA entries are three-seed means
(Appendix~\ref{app:simulation_seed_counts}). Dashes indicate no reported
result for the listed setting. Shades rank the five highest distinct scores
per column, from green to yellow. For $\pi_{0.5}$, LIBERO results follow the
\href{https://github.com/Physical-Intelligence/openpi/blob/main/examples/libero/README.md}{official openpi release};
SimplerEnv results follow \citet{lian2026langforce}.}
\label{tab:main_results}
\footnotesize
\setlength{\tabcolsep}{2.5pt}
\renewcommand{\arraystretch}{1.12}
\resizebox{\linewidth}{!}{%
\begin{tabular}{l*{10}{c}}
\toprule
\multirow{2}{*}{\textbf{Method}} &
\multicolumn{5}{c}{\textbf{LIBERO}} &
\multicolumn{5}{c}{\textbf{SimplerEnv-WidowX}} \\
\cmidrule(lr){2-6} \cmidrule(lr){7-11}
& Spatial & Object & Goal & Long & Avg.
& Spoon & Carrot & Stack & Eggplant & Avg. \\
\midrule
\multicolumn{11}{l}{\textit{Published results}} \\

OpenVLA~\citep{kim2024openvla}
& 84.7 & 88.4 & 79.2 & 53.7 & 76.5 & 0.0 & 0.0 & 0.0 & 4.1 & 1.0 \\

OpenVLA-OFT~\citep{kim2024openvla,kim2025fine}
& 97.6 & 98.4 & \topfour{97.9} & 94.5 & 97.1 & 12.5 & 4.2 & 8.3 & 37.5 & 15.6 \\

Octo-Small~\citep{octo_2023}
& 78.9 & 85.7 & 84.6 & 51.1 & 75.1 & 47.2 & 9.7 & 4.2 & 56.9 & 29.5 \\

$\pi_0$~\citep{black2024pi0}
& 96.8 & \topfour{98.8} & 95.8 & 85.2 & 94.2 & 29.1 & 0.0 & 16.7 & 62.5 & 27.1 \\

LoLA~\citep{wang2025lola}
& \topone{99.6} & \toptwo{99.6} & 97.2 & 88.2 & 96.2 & \toptwo{95.8} & 58.3 & \topfive{54.2} & 79.2 & 71.9 \\

LaMP~\citep{wang2026lamp}
& \toptwo{99.4} & \topone{99.8} & 97.4 & \topthree{96.7} & \toptwo{98.3} & 79.1 & \topfour{66.7} & \topthree{75.0} & \topthree{95.8} & \topthree{79.2} \\

UniVLA~\citep{wang2025univla}
& 95.4 & \topfour{98.8} & 93.6 & 94.0 & 95.5 & \topfive{83.3} & \topfour{66.7} & 33.3 & \topthree{95.8} & 69.8 \\

FlowVLA~\citep{zhong2025flowvla}
& 93.2 & 95.0 & 91.6 & 72.6 & 88.1 & 70.8 & 62.5 & \topfour{62.5} & \topone{100.0} & \topfive{74.0} \\

VLA-JEPA~\citep{sun2026vlajepa}
& 96.2 & \toptwo{99.6} & 97.2 & \topfive{95.8} & \topfive{97.2} & 75.0 & \toptwo{70.8} & 12.5 & 70.8 & 57.3 \\

DreamVLA~\citep{dreamvla25}
& 97.6 & \topfour{98.8} & 97.2 & 95.0 & \topfive{97.2} & 79.2 & 41.7 & 20.8 & \topone{100.0} & 60.4 \\

CoWVLA~\citep{yang2026cowvla}
& 97.2 & 97.8 & 94.6 & 92.8 & 95.6 & 79.2 & \topfour{66.7} & \topfour{62.5} & \topthree{95.8} & \topfour{76.0} \\

ThinkAct~\citep{huang2025thinkact}
& 88.3 & 91.4 & 87.1 & 70.9 & 84.4 & 58.3 & 37.5 & 8.7 & 70.8 & 43.8 \\

F1~\citep{lv2025f1}
& \topfive{98.2} & 97.8 & 95.4 & 91.3 & 95.7 & 50.0 & \toptwo{70.8} & 50.0 & 66.7 & 59.4 \\

UD-VLA~\citep{chen2025udvla}
& 94.1 & 95.7 & 91.2 & 89.6 & 92.7 & 58.3 & 62.5 & 54.1 & 75.0 & 62.5 \\

Fast-ThinkAct~\citep{huang2026fastthinkact}
& 92.0 & 97.2 & 90.2 & 79.4 & 89.7 & --- & --- & --- & --- & --- \\

LaRA-VLA~\citep{bai2026laravla}
& 96.4 & \topone{99.8} & \topone{98.6} & \topfour{96.6} & \topfour{97.9} & \toptwo{95.8} & 62.5 & 25.0 & \topfour{91.7} & 68.8 \\

X-VLA~\citep{zheng2025xvla}
& \topfive{98.2} & \topfive{98.6} & \topfive{97.8} & \topone{97.6} & \topthree{98.1} & \topone{100.0} & \topone{91.7} & \topone{95.8} & \topthree{95.8} & \topone{95.8} \\

$\pi_{0.5}$~\citep{physicalintelligence2025pi05}
& \topfour{98.8} & 98.2 & \topthree{98.0} & 92.4 & 96.9 & 49.3 & \topfive{64.7} & 44.7 & 69.7 & 57.1 \\

\midrule
\multicolumn{11}{l}{\textit{Shared-backbone comparisons}} \\

Qwen3-VL-FAST
& 97.3 & 97.4 & 96.3 & 90.6 & 95.4 & 18.8 & 31.3 & 4.2 & 71.9 & 31.6 \\

Qwen3-VL-OFT
& 97.8 & \topfive{98.6} & 96.2 & 93.8 & 96.6 & \topfour{90.3} & 38.5 & 29.7 & \topone{100.0} & 64.6 \\

Qwen3-VL-PI
& \topfour{98.8} & \toptwo{99.6} & 95.8 & 88.4 & 95.7 & 78.1 & 46.9 & 30.2 & \topfive{88.5} & 60.9 \\

Qwen3-VL-GR00T
& 97.8 & \topfour{98.8} & 97.4 & 92.0 & 96.5 & 83.0 & 59.4 & 18.8 & \topone{100.0} & 65.3 \\

\midrule

\textbf{SCULPT-VLA}
& \topthree{99.1} & \topthree{99.5} & \toptwo{98.3} & \toptwo{96.9} & \topone{98.5} & \topthree{90.7} & \topthree{70.7} & \toptwo{75.3} & \toptwo{97.3} & \toptwo{83.5} \\
\bottomrule
\end{tabular}}
\end{table}

% Main Table 2: published baselines and shared-backbone LIBERO / RoboTwin comparisons.
% Existing numbers preserved; shared-backbone LIBERO values repeat Table 1.
\begin{table}[!htbp]
\centering
\caption{Comparison on LIBERO and RoboTwin 2.0 Full (success, \%).
Published results retain their source protocols
(Appendix~\ref{app:baseline_parity}). Shared-backbone variants use Qwen3-VL
in StarVLA~\citep{community2026starvla}. RoboTwin Avg. averages Clean and Random.
$\dagger$: LIBERO reproduction by \citet{sun2026vladrop}; $\ddagger$: LIBERO
episode limits of 600 steps (Spatial/Object/Goal) and 700 (Long).
Motus's LIBERO and $\pi_0$'s RoboTwin results follow \citet{yuan2026fastwam}.
Darker and lighter green mark the best and second-best
distinct scores per column, respectively.}
\label{tab:libero_robotwin}
\footnotesize
\setlength{\tabcolsep}{3pt}
\renewcommand{\arraystretch}{1.15}
\resizebox{\linewidth}{!}{%
\begin{tabular}{l*{8}{c}}
\toprule
\multirow{2}{*}{\textbf{Method}} &
\multicolumn{5}{c}{\textbf{LIBERO}} &
\multicolumn{3}{c}{\textbf{RoboTwin 2.0 Full}} \\
\cmidrule(lr){2-6} \cmidrule(lr){7-9}
& Spatial & Object & Goal & Long & Avg.
& Clean & Random & Avg. \\
\midrule
\multicolumn{9}{l}{\textit{Published results}} \\
$\pi_0$~\citep{black2024pi0}
& 96.8 & 98.8 & 95.8 & 85.2 & 94.2 & 65.9 & 58.4 & 62.2 \\

LingBot-VLA~\citep{wu2026lingbotvla}
& 81.8$^{\dagger}$ & 95.0$^{\dagger}$ & 86.6$^{\dagger}$ & 67.8$^{\dagger}$ & 82.8$^{\dagger}$ & 88.6 & 86.7 & 87.7 \\

Motus~\citep{bi2025motus}
& 96.8 & \toptwo{99.8} & 96.6 & \toptwo{97.6} & 97.7 & 88.7 & 87.0 & 87.9 \\

Fast-WAM~\citep{yuan2026fastwam}
& 98.2 & \topone{100.0} & 97.0 & 95.2 & 97.6 & 91.9 & 91.8 & 91.9 \\

OpenWAM-$\alpha$~\citep{wang2026openwam}
& \topone{99.6} & 99.6 & \topone{99.8} & \topone{98.2} & \topone{99.3}$^{\ddagger}$ & \topone{93.7} & \topone{93.5} & \topone{93.6} \\

\midrule
\multicolumn{9}{l}{\textit{Shared-backbone comparisons}} \\

Qwen3-VL-FAST
& 97.3 & 97.4 & 96.3 & 90.6 & 95.4 & 72.5 & 83.2 & 77.9 \\

Qwen3-VL-OFT
& 97.8 & 98.6 & 96.2 & 93.8 & 96.6 & 88.2 & 88.3 & 88.3 \\

Qwen3-VL-PI
& 98.8 & 99.6 & 95.8 & 88.4 & 95.7 & 88.1 & 88.8 & 88.5 \\

Qwen3-VL-GR00T
& 97.8 & 98.8 & 97.4 & 92.0 & 96.5 & 88.0 & 88.5 & 88.3 \\

\midrule

\textbf{SCULPT-VLA}
& \toptwo{99.1} & 99.5 & \toptwo{98.3} & 96.9 & \toptwo{98.5} & \toptwo{92.3} & \toptwo{92.1} & \toptwo{92.2} \\
\bottomrule
\end{tabular}}
\end{table}

\begin{figure}[!htbp]
\centering
\includegraphics[width=\linewidth]{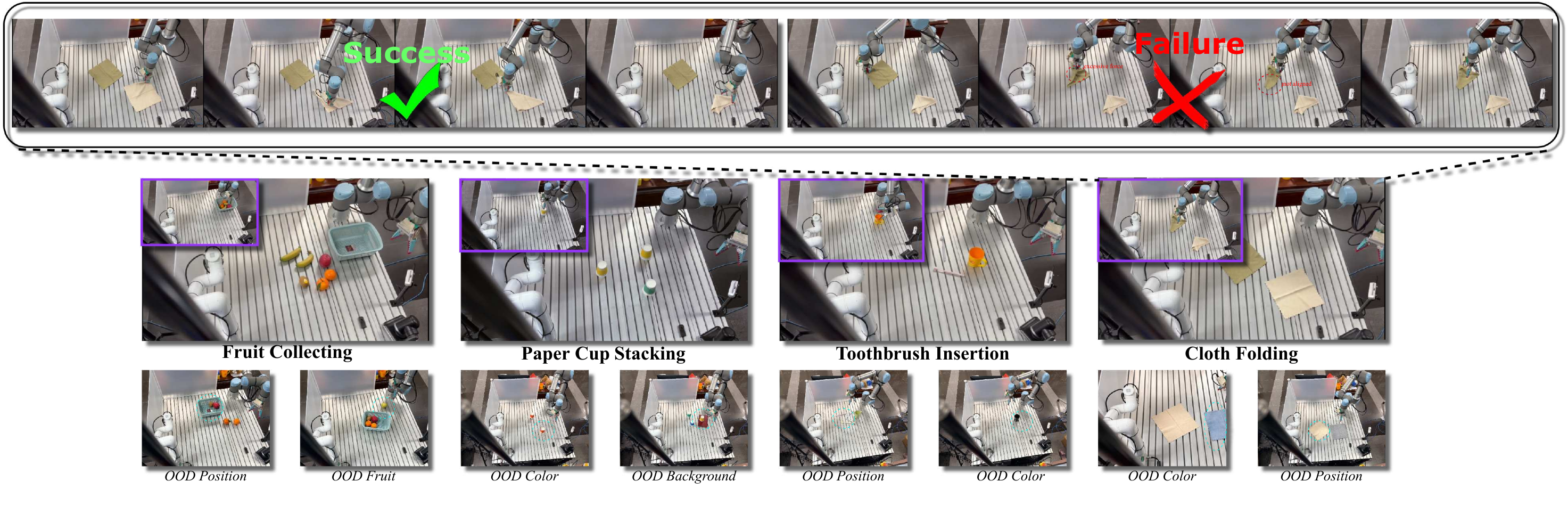}
\caption{Real-world evaluation on Fruit Collecting, Paper Cup Stacking,
Toothbrush Insertion, and Cloth Folding. Top: representative Cloth Folding success and failure. Middle: four
task setups. Bottom: task-specific OOD configurations.
Task definitions and perturbations are detailed in
Table~\ref{tab:realworld_ood_details}.}
\label{fig:realenv}
\end{figure}

\paragraph{Physical evaluation under distribution shift.}
\label{sec:exp-real}
The four physical tasks require object collection, stacking, insertion,
and deformable-object manipulation (Figure~\ref{fig:realenv}).
Each method is trained on 100 ID demonstrations per task.
We evaluate one trained checkpoint per method and task using three
independent evaluation batches of 30 physical rollouts per ID/OOD condition,
for 90 rollouts per condition.
Mean and sample standard deviation are computed across these three batches.
All compared methods follow the same task-specific ID/OOD evaluation protocol.
OOD settings change appearance, object properties, or placement relative
to these demonstrations (Appendix~\ref{app:real_world_setup}).
SCULPT-VLA achieves higher OOD success on all four tasks, averaging 58.1\%
versus 45.6\% for $\pi_{0.5}$ (Table~\ref{tab:real_world}).
Cloth Folding remains the hardest setting at 37.8\% OOD success.
Here, SCULPT-VLA's mOLS is 0.227 versus 0.229 for $\pi_{0.5}$.
Closed-loop success and open-loop action agreement thus rank the two
policies differently on this task.

\begin{table}[!htbp]
\centering
\caption{Real-world task success (\%). Each method uses 100 ID demonstrations
and one trained checkpoint per task. Entries are mean $\pm$ sample SD across
three independent evaluation batches of 30 physical rollouts per condition.
OOD settings are defined in
Table~\ref{tab:realworld_ood_details}; all open-loop scores are in
Table~\ref{tab:real_world_full}.}
\label{tab:real_world}
\footnotesize
\setlength{\tabcolsep}{2.8pt}
\renewcommand{\arraystretch}{1.1}
\begin{tabular}{lcccccccc}
\toprule
& \multicolumn{2}{c}{Fruit Collecting}
& \multicolumn{2}{c}{Cup Stacking}
& \multicolumn{2}{c}{Toothbrush Insertion}
& \multicolumn{2}{c}{Cloth Folding} \\
\cmidrule(lr){2-3}\cmidrule(lr){4-5}\cmidrule(lr){6-7}\cmidrule(lr){8-9}
Method & ID & OOD & ID & OOD & ID & OOD & ID & OOD \\
\midrule
OpenVLA-OFT & $50.0{\pm}3.3$ & $24.4{\pm}5.1$ & $61.1{\pm}1.9$ & $36.7{\pm}6.7$ & $44.4{\pm}1.9$ & $17.8{\pm}5.1$ & $25.6{\pm}1.9$ & $2.2{\pm}1.9$ \\
$\pi_0$ & $65.6{\pm}1.9$ & $45.6{\pm}6.9$ & $75.6{\pm}1.9$ & $56.7{\pm}3.3$ & $67.8{\pm}3.8$ & $44.4{\pm}5.1$ & $35.6{\pm}1.9$ & $14.4{\pm}3.8$ \\
$\pi_{0.5}$ & $74.4{\pm}1.9$ & $52.2{\pm}5.1$ & $81.1{\pm}1.9$ & $60.0{\pm}6.7$ & $73.3{\pm}3.3$ & $47.8{\pm}3.8$ & $41.1{\pm}1.9$ & $22.2{\pm}5.1$ \\
\textbf{SCULPT-VLA} & $80.0{\pm}3.3$ & $60.0{\pm}3.3$ & $85.6{\pm}1.9$ & $70.0{\pm}6.7$ & $84.4{\pm}5.1$ & $64.4{\pm}5.1$ & $56.7{\pm}3.3$ & $37.8{\pm}5.1$ \\
\bottomrule
\end{tabular}
\end{table}

\FloatBarrier
\subsection{Structured Organization and Action Grounding}
\label{sec:exp-ablation}
Table~\ref{tab:analysis_ablation} tests the representation and curriculum
on SimplerEnv-WidowX, including individual factor removals.
Variant definitions and per-threshold open-loop scores are in
Appendix~\ref{app:ablation_details}.
\begin{table}[!htbp]
\centering
\caption{Complete ablations on SimplerEnv-WidowX. FAST and Continuous denote the Stage-II discrete and Stage-III continuous policies. Success (\%) is reported as mean $\pm$ sample SD over evaluation runs; variants are trained independently. mOLS is defined in Appendix~\ref{app:open-loop}; per-threshold scores and variant definitions are in Appendix~\ref{app:ablation_details}.}
\label{tab:analysis_ablation}
\small
\setlength{\tabcolsep}{4pt}
\renewcommand{\arraystretch}{1.1}
\begin{tabular*}{0.94\linewidth}{@{\extracolsep{\fill}}lccc@{}}
\toprule
Variant & FAST & Continuous & mOLS $\uparrow$ \\
\midrule
\textbf{Full SCULPT-VLA} & $74.7{\pm}1.0$ & $83.5{\pm}0.0$ & 0.369 \\
\addlinespace[3pt]
shared factor FFN & $65.7{\pm}1.0$ & $76.8{\pm}1.0$ & 0.329 \\
generic latent pool & $52.8{\pm}1.2$ & $63.9{\pm}1.1$ & 0.271 \\
w/o subtask factor & $67.3{\pm}1.3$ & $78.3{\pm}1.3$ & 0.342 \\
w/o spatial factor & $64.8{\pm}0.8$ & $72.2{\pm}0.8$ & 0.312 \\
w/o dynamics factor & $61.6{\pm}1.0$ & $73.3{\pm}0.8$ & 0.322 \\
w/o action-side latent slots & $69.2{\pm}1.3$ & $77.0{\pm}1.0$ & 0.335 \\
\addlinespace[3pt]
w/o Stage-II visibility restriction & $55.7{\pm}0.8$ & $71.3{\pm}1.0$ & 0.302 \\
w/o Stage I & $60.0{\pm}1.0$ & $70.2{\pm}1.0$ & 0.301 \\
Stage III only & --- & $61.0{\pm}1.0$ & 0.262 \\
w/o FAST supervision & --- & $68.2{\pm}1.3$ & 0.290 \\
w/o DINOv2 detail branch & $73.7{\pm}1.0$ & $81.0{\pm}1.0$ & 0.354 \\
\bottomrule
\end{tabular*}
\end{table}

\paragraph{Structured organization.}
A generic latent pool achieves 63.9\% success, compared with 83.5\%
for the full model. This control preserves the teacher targets,
token budget, action objectives, and three-stage schedule, while jointly
removing explicit type identities, factor-specific FFNs, and type-dependent
visibility constraints. It tests their combined contribution.
Sharing the factor FFN gives 76.8\%, testing factor-specific computation.
Removing subtask, spatial, or dynamics factors tests the respective
factor families; each removal reduces both FAST and continuous success.
The largest factor-removal drop shifts from dynamics under FAST decoding
to spatial grounding under continuous control.
Removing action-side latent slots leaves factor states available for
decoding, but degrades both policies. This supports using a dedicated
readout to combine factor information for action prediction.

\paragraph{Action grounding.}
Allowing direct vision--language access to action positions in Stage~II
reduces success to 71.3\%.
The deficit is already present under FAST decoding and persists in
Stage~III, when both variants have the same direct perceptual access.
This supports establishing factor-conditioned action prediction before
continuous refinement. The benefit is task-dependent:
unrestricted Stage-II access gives
98.7\% on LIBERO-Spatial, close to the standard 99.1\%
(Appendix~\ref{app:stage3_shortcut}).
Removing FAST supervision tests the coarse action objective
and yields 68.2\% success.

\paragraph{Training stages and refinement.}
Skipping Stage~I yields 70.2\%, and the Stage III only variant gives 61.0\%.
The latter retains the SCULPT-VLA architecture and auxiliary losses,
distinguishing it from the direct Qwen3-VL-PI baseline.
Omitting training stages also changes the total optimization budget
under these schedules.
Within Stage~III, removing the DINOv2 detail branch gives 81.0\%,
a smaller reduction than removing structured organization.

Across the reported variants, mOLS and mean continuous success rank the
models in the same order. The gains therefore appear in both action
prediction on demonstration observations and closed-loop task execution.
The following analysis complements these training-time ablations with
interventions on a fixed trained policy, testing whether the resulting
factors remain functionally involved during continuous control.

% Keep the half-width intervention table with its explanatory paragraph.
\clearpage
\subsection{Functional Analysis of Learned Structured Control States}
\label{sec:exp-qual}
\begin{wraptable}{r}{0.50\linewidth}
\vspace{-\baselineskip}
\centering
\captionsetup{font=small,skip=5pt}
\caption{Factor-wise intervention on Fruit Collecting. Selected final factor states are replaced with their training-set means before continuous decoding. Success (\%): mean $\pm$ sample SD across three evaluation batches of 30 rollouts per condition. Raw counts: Table~\ref{tab:slot_replacement}.}
\label{tab:slot_replacement_main}
\small
\setlength{\tabcolsep}{4pt}
\renewcommand{\arraystretch}{1.1}
\begin{tabular}{lcc}
\toprule
Replaced states & ID & OOD \\
\midrule
Full SCULPT-VLA & $80.0{\pm}3.3$ & $60.0{\pm}3.3$ \\
Replace subtask factor & $54.4{\pm}5.1$ & $28.9{\pm}10.2$ \\
Replace spatial factor & $73.3{\pm}3.3$ & $50.0{\pm}6.7$ \\
Replace dynamics factor & $65.6{\pm}5.1$ & $42.2{\pm}6.9$ \\
Replace all factors & $15.6{\pm}5.1$ & $5.6{\pm}5.1$ \\
\bottomrule
\end{tabular}
\end{wraptable}

\paragraph{Factor-wise functional intervention.}
We replaced selected final factor states with their training-set means
before continuous decoding, retaining the backbone, action readout
states, and DINOv2 features.
On Fruit Collecting, replacing all three factors reduced success from
80.0\% to 15.6\% ID and from 60.0\% to 5.6\% OOD
(Table~\ref{tab:slot_replacement_main}).
Subtask replacement had the largest effect among individual groups.
These behavioral changes show that the learned structured state
contributes to deployed control after direct perceptual access is restored.
The intervention protocol is described in Appendix~\ref{app:slot_intervention}.

\begin{figure}[!htbp]
  \centering
  \includegraphics[width=\linewidth]{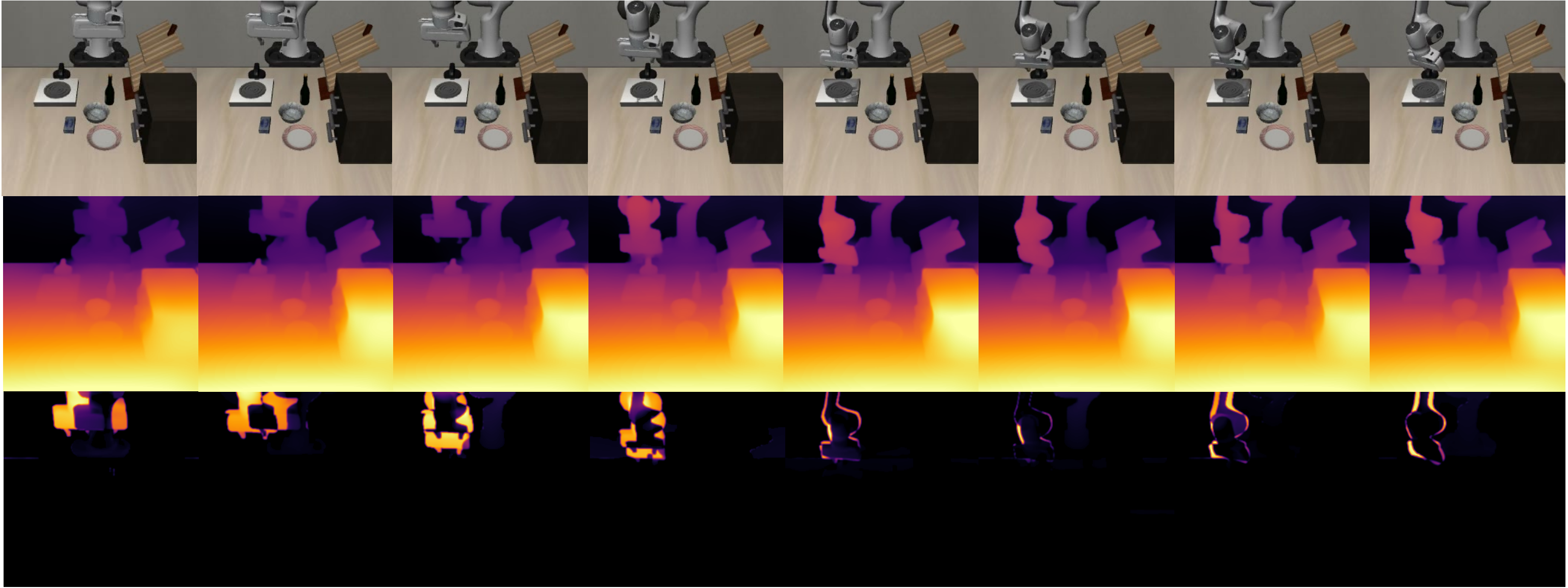}
  \caption{Factor probes on \textit{turn\_on\_the\_stove}.
  Rows show observations, depth reconstructed from spatial slots,
  and depth-change predictions from dynamics slots.
  These probes visualize decodable information;
  Appendix~\ref{app:spa_dyn_probe} describes the diagnostic setup.}
  \label{fig:probe_main}
\end{figure}

\paragraph{Decodable factor content.}
Figure~\ref{fig:probe_main} visualizes depth reconstructed from spatial
factors and depth change predicted from dynamics factors.
The separately trained probes reveal scene geometry and state-transition
cues in the learned state, complementing the behavioral
interventions.

Further representation, attention, and trajectory analyses are reported
in Appendix~\ref{app:internalization_analysis}.

\FloatBarrier
\section{Conclusion and Limitations}
SCULPT-VLA connects structured supervision to control through staged
action grounding. Closed-loop evaluations on LIBERO, SimplerEnv-WidowX,
RoboTwin 2.0 Full, and four physical robot tasks support this design.
Training controls and factor-wise functional interventions show the
learned state's contribution after direct perceptual access is restored.
The results suggest that learning how to use structure matters
alongside learning what that structure encodes.

The current design uses fixed control factors and training-time
teachers. Physical evaluation covers four tabletop tasks on one UR5e platform
(Appendix~\ref{app:limitations}).

\clearpage
\label{page:statements-start}
\section*{AI Disclosure}
Generative AI tools were used for language polishing of selected sentences.
All scientific content, analyses, and reported results were reviewed and
verified by the authors. Qwen3-VL-32B-Instruct was used in the automated
subtask-annotation pipeline described in Appendix~\ref{sec:video2tasks_pipeline}.

\section*{Reproducibility Statement}
Section~\ref{sec:training} describes staged action grounding.
Appendix~\ref{sec:appendix_training_details} provides training and
subtask-annotation details.
Appendix~\ref{app:exp-setup} gives evaluation protocols, ablation
definitions, and complete results.
Appendix~\ref{app:internalization_analysis} describes the probes and
factor-wise intervention procedure.

\bibliographystyle{iclr2027_conference}
\bibliography{refs}
\clearpage
\appendix
\label{page:appendix-start}
\makeatletter
\setlength{\@fptop}{0pt}
\setlength{\@fpsep}{10pt}
\setlength{\@fpbot}{0pt plus 1fil}
\makeatother
\raggedbottom
\section*{Appendix}
This appendix provides the training configuration, evaluation protocols
and complete results, functional and representation analyses,
additional rollouts, and the scope of the method.

\begin{center}
\small
\renewcommand{\arraystretch}{1.12}
\begin{tabularx}{0.98\linewidth}{lXr}
\toprule
& Contents & Page \\
\midrule
\ref{sec:appendix_training_details} & \hyperref[sec:appendix_training_details]{Training and data construction} & \pageref{sec:appendix_training_details} \\
\ref{app:eval_details} & \hyperref[app:eval_details]{Evaluation protocols and complete results} & \pageref{app:eval_details} \\
\ref{app:internalization_analysis} & \hyperref[app:internalization_analysis]{Functional and representation analysis} & \pageref{app:internalization_analysis} \\
\ref{app:qualitative_rollouts} & \hyperref[app:qualitative_rollouts]{Additional rollouts} & \pageref{app:qualitative_rollouts} \\
\ref{app:limitations} & \hyperref[app:limitations]{Design scope and limitations} & \pageref{app:limitations} \\
\bottomrule
\end{tabularx}
\end{center}

\section{Training and Data Construction}
\label{sec:appendix_training_details}
\subsection{Teacher Scaffolds and Token Layout}

\paragraph{Teacher inputs and alignment targets.}
The frozen teachers are Qwen3-Embedding for task semantics
\citep{zhang2025qwen3}, UniVLA for task-relevant inverse-dynamics cues
\citep{bu2025univla}, and Depth Anything V2 for current-scene geometry
\citep{yang2024depth}. The corresponding slot counts are $4$ for
subtask, $4$ for dynamics, and $8$ for spatial representations.
All benchmarks and real-robot tasks use these teacher sources and
the same three-stage training design.
\label{sec:appendix_scaffold_init}

For each factor $\tau$, the trainable input projector produces
$\tilde{\mathbf h}_t^\tau=P_{\mathrm{in}}^\tau(\mathbf f_t^\tau)$
from a detached teacher feature.
This scaffold supplies the slot input before annealing.
The alignment reference is separately detached,
$\bar{\mathbf h}_t^\tau=\mathrm{sg}(\tilde{\mathbf h}_t^\tau)$,
so the target branch updates neither the teacher nor the input projector.
Input projectors are trained through scaffold inputs and output
projectors through alignment; the teacher encoders remain frozen.
The projectors/adapters in Table~\ref{tab:freezing_strategy} comprise
teacher-to-slot input projections, student-side alignment projections,
and slot-adaptation modules.

\paragraph{Training and deployment sequences.}
The formats below instantiate the three objectives in the main Method.
All stages include visual and instruction tokens. Stage~I adds the
three factor groups; Stages~II and III also use four fixed action
readout slots. Stage~II appends teacher-forced FAST positions for
discrete supervision and anneals scaffold inputs into learned queries.
Stage~III continuous-control inputs contain factor and fixed readout
slots, without teacher-forced FAST tokens or autoregressive generation.
The continuous head conditions on these states and DINOv2 detail features.
Weak FAST supervision uses a separate discrete auxiliary path;
alignment targets remain active. Deployment omits teacher encoders
and the auxiliary path.

\begin{figure*}[t]
\centering
\begin{tcolorbox}[
    width=0.97\textwidth,
    colback=white,
    colframe=black!55,
    boxrule=0.5pt,
    arc=3mm,
    left=3mm,right=3mm,top=2mm,bottom=2mm
]
\small
\ttfamily
\raggedright

{\color{stageblue}\bfseries Stage I: Latent Alignment}\par
Task: ``pick up the potato and place it into the bowl''\par
Input:\par
Your task is to pick up the potato and place it into the bowl.\par
\tok{SOST} \slotrep{slot_subtask}{4} \tok{EOST}\par
\tok{SOSD} \slotrep{slot_dynamic}{4} \tok{EOSD}\par
\tok{SOSS} \slotrep{slot_spatial}{8} \tok{EOSS}\par
Main supervision:\par
Align \tok{slot_subtask}, \tok{slot_dynamic},\par and \tok{slot_spatial} to frozen teacher targets under \(\mathbf{M}^{(1)}\).\par
{\color{notegray}Note: scaffold-derived slot inputs are fully used;\par coarse-action supervision and the act-typed group are not active.}\par

\vspace{0.6em}

{\color{stageblue}\bfseries Stage II: Scaffold-Input Annealing with Coarse Action Grounding}\par
Task: ``open the drawer and place the sponge inside''\par
Input:\par
Your task is to open the drawer and place the sponge inside.\par
\tok{SOST} \slotrep{slot_subtask}{4} \tok{EOST}\par
\tok{SOSD} \slotrep{slot_dynamic}{4} \tok{EOSD}\par
\tok{SOSS} \slotrep{slot_spatial}{8} \tok{EOSS}\par
\tok{SOSA} \slotrep{slot_action}{4} \tok{EOSA}\par
\tok{SOFA} \slotrep{fast_action}{M} \tok{EOFA}\par
Main supervision:\par
FAST next-token loss at act-typed FAST positions,\par with continuing teacher alignment for \tok{slot_subtask},\par \tok{slot_dynamic}, and \tok{slot_spatial} under \(\mathbf{M}^{(2)}\).\par
{\color{notegray}Note: scaffold inputs to structured slots are annealed\par into learned query embeddings; output-side\par teacher alignment remains active throughout.}\par

\vspace{0.6em}

{\color{stageblue}\bfseries Stage III: Continuous Action Refinement}\par
Task: ``push the cup to the target mark''\par
Continuous-policy input:\par
Your task is to push the cup to the target mark.\par
\tok{SOST} \slotrep{slot_subtask}{4} \tok{EOST}\par
\tok{SOSD} \slotrep{slot_dynamic}{4} \tok{EOSD}\par
\tok{SOSS} \slotrep{slot_spatial}{8} \tok{EOSS}\par
\tok{SOSA} \slotrep{slot_action}{4} \tok{EOSA}\par
Main supervision:\par
Continuous action target via flow matching under \(\mathbf{M}^{(3)}\).\par
{\color{notegray}Note: weak FAST supervision uses a separate\par autoregressive path, outside this continuous-policy sequence.\par The continuous head uses factor states, fixed action readout,\par and DINOv2 detail features.}\par

\end{tcolorbox}
\caption{\textbf{Example training formats across the three stages of
SCULPT-VLA.}
\texttt{<slot\_subtask>}, \texttt{<slot\_dynamic>},
\texttt{<slot\_spatial>}, and \texttt{<slot\_action>} denote
placeholder tokens reserved in the VLM sequence for subtask, dynamics,
spatial, and action-side latent slots, respectively. Stage~II
additionally appends teacher-forced FAST action-token positions,
assigned type \texttt{act}, for coarse next-token supervision.
The Stage~III continuous sequence excludes FAST tokens; weak FAST
supervision uses a separate autoregressive path.
The continuous head reads final factor and action-readout states,
together with DINOv2 detail features. Boundary tokens such as \texttt{<SOST>} /
\texttt{<EOST>} mark the start and end of each slot block for type
assignment, masking, and stage-specific loss computation.}
\end{figure*}

\subsection{Optimization and Trainability}
\label{app:optimization}

\paragraph{Training configuration.}
Within each benchmark, the action horizon is
fixed across the three stages. Unless otherwise specified, all stages
use AdamW with a cosine learning-rate scheduler and warm-up ratio
$0.1$. Detailed settings are reported in
Table~\ref{tab:training_hparams}.

\paragraph{Parameter adaptation.}
Table~\ref{tab:freezing_strategy} specifies module trainability.
Stage~I updates slot-specific modules.
Stage~II adds LoRA updates to selected VLM attention projections and
trains the action-related modules.
Stage~III trains the continuous head while co-training the previously
learned pathway at the listed reduced learning rates.
The base FFN and teacher encoders stay frozen.
LoRA applies only to selected VLM linear projections;
slot projectors, typed FFNs, condition fusion, and the continuous
DiT head use full-rank optimization when active.

\begin{table*}[t]
    \centering
    \caption{Stage-wise module trainability in SCULPT-VLA.
    \trainable~denotes trainable modules, \frozenm~denotes frozen
    modules, and \lowlr~denotes modules co-trained with a reduced
    learning rate. For the VLM backbone, ``trainable'' denotes
    LoRA-updated linear projections rather than full-rank backbone
    training.}
    \label{tab:freezing_strategy}
    \scriptsize
    \setlength{\tabcolsep}{4.4pt}
    \renewcommand{\arraystretch}{1.10}
    \resizebox{\textwidth}{!}{%
    \begin{tabular}{lccccccccc}
        \toprule
        \textbf{Stage}
        & \shortstack{\textbf{Teacher}\\\textbf{Encoders}}
        & \shortstack{\textbf{VLM Attn}\\\textbf{LoRA}}
        & \shortstack{\textbf{Base-}\\\textbf{FFN}}
        & \shortstack{\textbf{Sub./Dyn./Spa.}\\\textbf{FFNs}}
        & \shortstack{\textbf{Action}\\\textbf{FFN}}
        & \shortstack{\textbf{Slot Proj./}\\\textbf{Adapters}}
        & \shortstack{\textbf{FAST Action}\\\textbf{LM Head}}
        & \shortstack{\textbf{DiT Action}\\\textbf{Head}} \\
        \midrule
        Stage I
        & \frozenm
        & \frozenm
        & \frozenm
        & \trainable
        & \frozenm
        & \trainable
        & \frozenm
        & \frozenm \\

        Stage II
        & \frozenm
        & \trainable
        & \frozenm
        & \trainable
        & \trainable
        & \trainable
        & \trainable
        & \frozenm \\

        Stage III
        & \frozenm
        & \lowlr
        & \frozenm
        & \lowlr
        & \lowlr
        & \trainable
        & \lowlr
        & \trainable \\
        \bottomrule
    \end{tabular}}
\end{table*}

\begin{table}[t]
  \centering
  \caption{Hyperparameter settings of SCULPT-VLA across LIBERO,
  RoboTwin 2.0, SimplerEnv, and real-robot experiments. \emph{Scheduled} indicates
  scaffold-input annealing or coarse-loss weighting following the
  schedule in Sec.~\ref{sec:training}; \emph{Weak} denotes
  retained at a small weight as auxiliary regularization.}
  \label{tab:training_hparams}
  \scriptsize
  \setlength{\tabcolsep}{5pt}
  \renewcommand{\arraystretch}{1.12}
  \resizebox{\textwidth}{!}{%
  \begin{tabular}{lccccccccc}
    \toprule
    \multirow{2}{*}{\textbf{Hyperparameters}}
    & \multicolumn{3}{c}{\textbf{LIBERO / RoboTwin 2.0}}
    & \multicolumn{3}{c}{\textbf{SimplerEnv}}
    & \multicolumn{3}{c}{\textbf{Real Robot}} \\
    \cmidrule(lr){2-4} \cmidrule(lr){5-7} \cmidrule(lr){8-10}
    & \textbf{Stage I} & \textbf{Stage II} & \textbf{Stage III}
    & \textbf{Stage I} & \textbf{Stage II} & \textbf{Stage III}
    & \textbf{Stage I} & \textbf{Stage II} & \textbf{Stage III} \\
    \midrule

    \multicolumn{10}{l}{\textit{Learning Rates}} \\
    VLM LR
      & --- & $1\times10^{-5}$ & $5\times10^{-6}$
      & --- & $1\times10^{-5}$ & $5\times10^{-6}$
      & --- & $1\times10^{-5}$ & $5\times10^{-6}$ \\
    Teacher Encoder LR
      & --- & --- & ---
      & --- & --- & ---
      & --- & --- & --- \\
    Slot Projector LR
      & $1\times10^{-4}$ & $1\times10^{-4}$ & $5\times10^{-5}$
      & $1\times10^{-4}$ & $1\times10^{-4}$ & $5\times10^{-5}$
      & $1\times10^{-4}$ & $1\times10^{-4}$ & $5\times10^{-5}$ \\
    FAST Action Head LR
      & --- & $1\times10^{-4}$ & $5\times10^{-5}$
      & --- & $1\times10^{-4}$ & $5\times10^{-5}$
      & --- & $1\times10^{-4}$ & $5\times10^{-5}$ \\
    Continuous Action Head LR
      & --- & --- & $1\times10^{-4}$
      & --- & --- & $1\times10^{-4}$
      & --- & --- & $1\times10^{-4}$ \\
    \midrule

    \multicolumn{10}{l}{\textit{Optimization Config}} \\
    Action Horizon
      & 8 & 8 & 8
      & 8 & 8 & 8
      & 16 & 16 & 16 \\
    Training Steps
      & 5k & 20k & 20k
      & 5k & 20k & 20k
      & 3k & 15k & 15k \\
    Batch Size
      & 8 & 4 & 8
      & 8 & 4 & 8
      & 8 & 4 & 8 \\
    Optimizer
      & AdamW & AdamW & AdamW
      & AdamW & AdamW & AdamW
      & AdamW & AdamW & AdamW \\
    LR Scheduler
      & Cosine & Cosine & Cosine
      & Cosine & Cosine & Cosine
      & Cosine & Cosine & Cosine \\
    Warm-up Ratio
      & 0.1 & 0.1 & 0.1
      & 0.1 & 0.1 & 0.1
      & 0.1 & 0.1 & 0.1 \\
    \midrule

    \multicolumn{10}{l}{\textit{Loss Weights}} \\
    Slot Alignment Loss (Sub./Dyn./Spa.)
      & $1.0/1.0/1.0$ & Scheduled & Weak
      & $1.0/1.0/1.0$ & Scheduled & Weak
      & $1.0/1.0/1.0$ & Scheduled & Weak \\
    Coarse Action Loss
      & --- & Scheduled & Weak
      & --- & Scheduled & Weak
      & --- & Scheduled & Weak \\
    Continuous FM Loss
      & --- & --- & Main
      & --- & --- & Main
      & --- & --- & Main \\
    Continuous Action Head
      & Off & Off & On
      & Off & Off & On
      & Off & Off & On \\
    \bottomrule
  \end{tabular}}
\end{table}

\subsection{Subtask Annotation Pipeline}
\label{sec:video2tasks_pipeline}

\begin{figure*}[t]
  \centering
  \includegraphics[width=\textwidth]{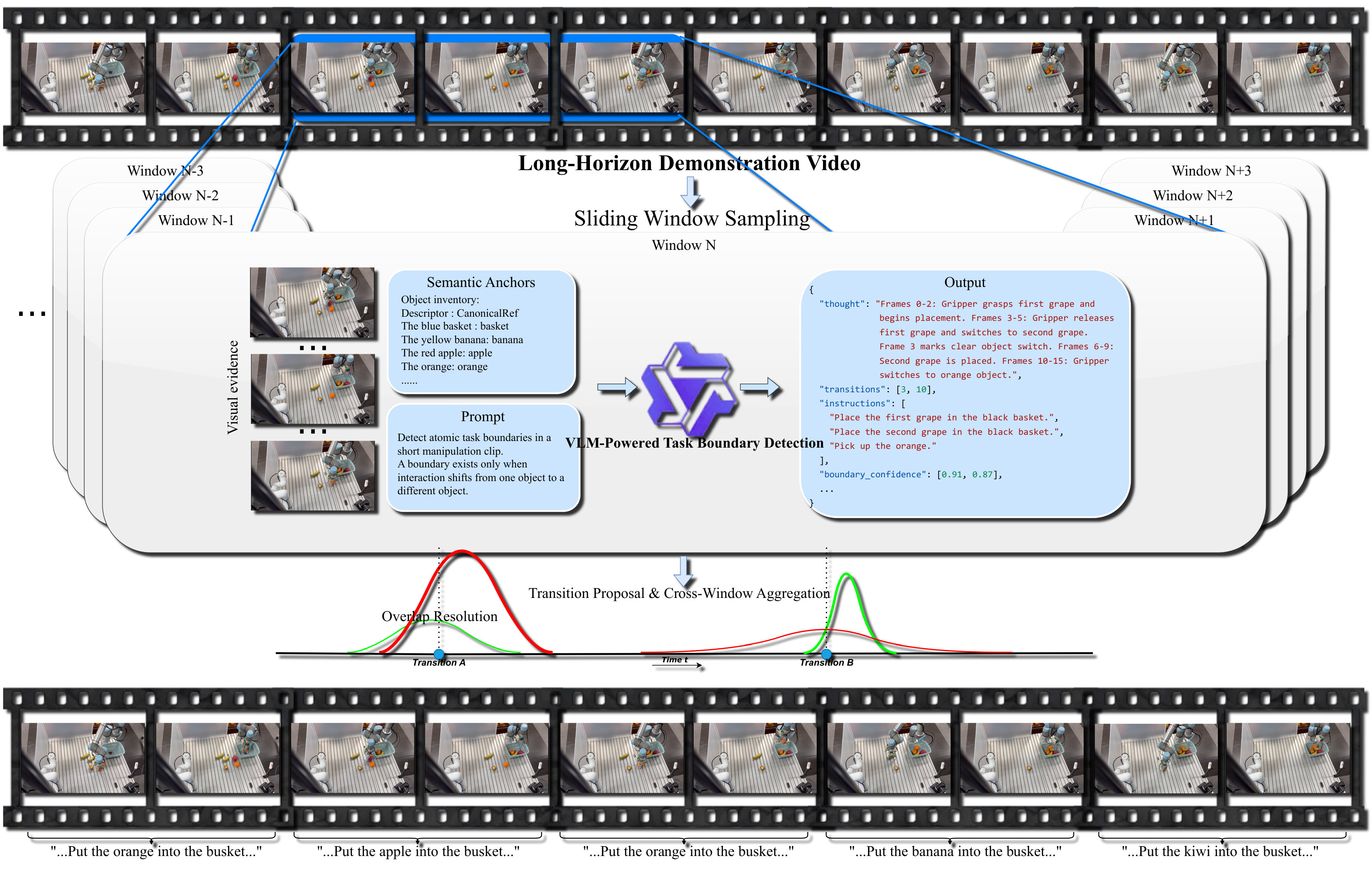}
  \caption{Automated long-video segmentation pipeline. An untrimmed
  demonstration video is decomposed into overlapping temporal
  windows. Within each window, sampled visual evidence and semantic
  anchors are provided to Qwen3-VL-32B-Instruct for object-centric
  transition prediction. The resulting transition proposals are mapped
  to the global timeline and merged through cross-window aggregation
  to remove redundant detections from overlapping windows. The final
  consolidated boundaries partition the original demonstration into
  atomic task segments with aligned language instructions.}
  \label{fig:video2tasks_pipeline}
\end{figure*}

\textbf{LIBERO-SUBTASK} and \textbf{BRIDGE-SUBTASK} are constructed
through an automated pipeline that decomposes an untrimmed
manipulation video into a sequence of temporally grounded atomic task
segments with aligned language instructions
(Fig.~\ref{fig:video2tasks_pipeline}). The pipeline comprises three
stages: window-wise transition prediction, global boundary
consolidation, and segment-level instruction assignment. The output
provides the subtask-stage signal used by Qwen3-Embedding when
generating instruction-aware features~\citep{zhang2025qwen3} for the subtask slots.
The same annotation procedure is applied to RoboTwin and real-robot
demonstrations to construct their task-progression scaffolds.
These training annotations are separate from the analysis-only phase
labels used in Appendix~\ref{app:subtask_pca}.

\paragraph{Implementation parameters.}
Unless otherwise specified, videos are processed at their native
frame rate with temporal windows of $64$ frames and stride $32$. For
each window, we uniformly sample $8$ frames and provide them to the
locally hosted Qwen3-VL-32B-Instruct model together with a compact
object-inventory prompt. The annotation model is constrained to return a JSON record with fields
\texttt{thought}, \texttt{transitions}, \texttt{instructions}, and
\texttt{boundary\_confidence}; these are mapped to the internal metadata
fields \texttt{rationale}, \texttt{transition\_frames}, and
\texttt{segment\_instructions} for downstream use.
Outputs that fail JSON parsing or contain non-monotone transition
indices are regenerated once and otherwise discarded. Cross-window
transition proposals within $12$ frames are merged by taking the
median proposal time, and segments shorter than $8$ frames after
consolidation are removed. Boundary confidence is the VLM-reported
confidence for each proposed transition after cross-window
consolidation. It serves as a filtering and quality-control score,
not a training target. We manually inspect all parse-failure cases
and a random $10\%$ subset of valid annotations.

\paragraph{Source scale.}
LIBERO-SUBTASK is derived from the four standard LIBERO suites used
in our evaluation, comprising $2{,}000$ source demonstrations in
total ($4$ suites $\times$ $10$ tasks $\times$ $50$
demonstrations)~\citep{liu2023libero}. BRIDGE-SUBTASK is derived from
the WidowX/BridgeData source pool used for SimplerEnv-style training;
public BridgeData V2 releases are on the order of
$5\times 10^{4}$--$6\times 10^{4}$ trajectories depending on filtering
and release version~\citep{walke2023bridgedata}.
For RoboTwin 2.0, we use the Full training set comprising 50 tasks,
with 50 clean and 500 randomized demonstrations per task, for 27,500
source demonstrations in total. The same annotation pipeline is
applied to these demonstrations to construct task-progression scaffolds.
Since the pipeline may split one source trajectory into multiple atomic segments and discard
invalid or very short proposals, we report source-trajectory scale
here and store the retained segment count and mean segment length in
the generated annotation metadata.

\paragraph{Window-wise transition prediction.}
Within each temporal window, the sampled frames are paired with
semantic anchors specifying the local
object inventory and canonical object references. These inputs are
provided to Qwen3-VL-32B-Instruct prompted to detect atomic task
transitions under the fixed JSON schema. We define a transition as a
switch in interaction from one manipulated object to another.
This criterion suppresses spurious boundaries from motion within a
single object-centric manipulation phase.

\paragraph{Global boundary consolidation.}
Local transition indices are mapped onto the global video timeline.
Overlapping windows often predict the same boundary with small temporal
offsets. We consolidate these proposals using the temporal merge
tolerance above, yielding a consistent segmentation of the demonstration.

\paragraph{Instruction assignment.}
The instruction candidates generated in local windows are aligned
with the consolidated temporal segments to assign one language
instruction to each atomic task interval. Candidate instructions that
disagree with the consolidated start/end interval are filtered during
the same metadata validation pass.

\subsection{Annotation Pipeline Statistics}
\label{app:subtask_quality}

Table~\ref{tab:subtask_quality} summarizes parsing validity, discarded
segments, and VLM-reported boundary confidence after consolidation.

\begin{table}[!htbp]
\centering
\caption{Processing statistics of the derived subtask annotations. Validity
and discard rates are computed after one regeneration attempt; confidence
denotes the percentage of valid annotations whose VLM-reported boundary
confidence exceeds $80\%$.}
\label{tab:subtask_quality}
\scriptsize
\setlength{\tabcolsep}{6pt}
\renewcommand{\arraystretch}{1.12}
\resizebox{0.78\textwidth}{!}{%
\begin{tabular}{lccc}
\toprule
\textbf{Dataset} &
\textbf{JSON valid} &
\textbf{Discarded} &
\textbf{Conf. $>80\%$} \\
\midrule
LIBERO-SUBTASK &
$98.1\%$ & $3.0\%$ & $94.8\%$ \\
BRIDGE-SUBTASK &
$96.2\%$ & $5.8\%$ & $91.5\%$ \\
\bottomrule
\end{tabular}}
\end{table}

\FloatBarrier
\section{Evaluation Protocols and Complete Results}
\label{app:eval_details}
\label{app:exp-setup}
\label{app:extended_results}
This section details the protocols for the main comparisons in
Tables~\ref{tab:main_results} and~\ref{tab:libero_robotwin}, followed by
simulation seed counts, ablation definitions, and real-world results.

\subsection{Comparison Scope and Evaluation Protocols}
\label{app:baseline_parity}
Published methods retain their source-reported protocols.
Shared-backbone controls use Qwen3-VL under the StarVLA evaluation stack,
with shared robot data, action horizon, and evaluation interface
\citep{community2026starvla,bai2025qwen3}.
They use FAST, OFT, $\pi_0$-style flow, or GR00T-style action heads
\citep{pertsch2025fast,kim2025fine,black2024pi0,nvidia2025gr00t}.
Protocol differences affecting direct comparability are noted in the
corresponding table captions, including the $\dagger$ and $\ddagger$
entries in Table~\ref{tab:libero_robotwin}.

SCULPT-VLA follows the RoboTwin 2.0 Full protocol
\citep{chen2025robotwin,community2026starvla}, with joint training over
50 tasks using 50 clean and 500 randomized demonstrations per task.
A single final checkpoint is evaluated on 100 Clean and 100 Random
episodes for each task. Clean and Random scores are arithmetic means
of the 50 task-level success rates; Avg. averages these two scores.
Both conditions are represented during training.
Table~\ref{tab:libero_robotwin} reports the main comparison, and
Table~\ref{tab:robotwin_additional} gives additional published results.
% Additional verified RoboTwin entries moved from the former wide main table.
\begin{table}[!htbp]
\centering
\caption{Additional published RoboTwin 2.0 Full results (success, \%).
X-VLA results follow \citet{bi2025motus}, and $\pi_{0.5}$ results follow
\citet{yuan2026fastwam}. Avg. is the arithmetic mean of Clean and Random.}
\label{tab:robotwin_additional}
\small
\setlength{\tabcolsep}{9pt}
\renewcommand{\arraystretch}{1.12}
\begin{tabular}{lccc}
\toprule
Method & Clean & Random & Avg. \\
\midrule
X-VLA~\citep{zheng2025xvla}
& 72.8 & 72.8 & 72.8 \\

$\pi_{0.5}$~\citep{physicalintelligence2025pi05}
& 82.7 & 76.8 & 79.8 \\

\bottomrule
\end{tabular}
\end{table}

\subsection{Open-Loop Evaluation Details}
\label{app:open-loop}

We additionally measure agreement with held-out demonstration action
chunks on SimplerEnv-WidowX \citep{li2024evaluating} and the real-world
ID/OOD validation splits. The latter are described in
Appendix~\ref{app:realworld_stats}.
Each prediction is compared with the corresponding expert action chunk.

Formally, let $N$ denote the number of evaluation transitions and let
$H$ denote the action horizon. For transition $i$, the model predicts
an action chunk $\hat{\mathbf{a}}_{i,0:H-1}$, which is compared with
the corresponding expert chunk $\mathbf{a}^{\ast}_{i,0:H-1}$. OLS is
computed on the normalized 6-DoF continuous action component only,
excluding the gripper dimension. For a raw 6-DoF action vector
$\mathbf{a}$, we define the normalization map
\[
\phi(\mathbf{a})_d
=
\frac{\mathbf{a}_d-\mu_d}{s_d+\epsilon},
\qquad
s_d=\mathrm{P}_{95}\!\left(\left|\mathbf{a}_d-\mu_d\right|\right),
\]
where $\mu_d$ and $s_d$ are computed from the training split for each
translation or rotation dimension, $\mathrm{P}_{95}$ denotes the
$95$th percentile, and $\epsilon=10^{-6}$. Translation and rotation
errors are therefore measured in the same normalized action space.
For threshold $\delta$,
\begin{equation}
\mathrm{OLS}@\delta
=
\frac{1}{N}\sum_{i=1}^{N}
\mathbf{1}\!\left(
\frac{1}{H}\sum_{k=0}^{H-1}
\left\lVert
\phi(\hat{\mathbf{a}}_{i,k})-\phi(\mathbf{a}^{\ast}_{i,k})
\right\rVert_2
< \delta
\right)
\label{eq:ols}
\end{equation}
A predicted chunk counts as correct when its average per-step
Euclidean error is below $\delta$ in normalized 6-DoF action space.
The thresholds in $\Delta$ are dimensionless.

We further report the mean open-loop score
\begin{equation}
\mathrm{mOLS}
=
\frac{1}{|\Delta|}
\sum_{\delta\in\Delta}\mathrm{OLS}@\delta,
\qquad
\Delta=\{0.1,\,0.05,\,0.03,\,0.01\}
\label{eq:mols}
\end{equation}
Higher OLS and mOLS indicate closer agreement with expert action chunks. Both
metrics are averaged over all sampled transitions in the held-out
split.

\subsection{Simulation Results and Seed Counts}
\label{app:simulation_seed_counts}
Table~\ref{tab:simulation_seed_counts} reports the seed-level closed-loop
success counts of SCULPT-VLA for the main simulation results on
LIBERO~\citep{liu2023libero} and
SimplerEnv-WidowX~\citep{li2024evaluating}. For LIBERO, each suite-level entry aggregates 10 tasks with 50 trials per
task, yielding 500 trials per seed. For SimplerEnv-WidowX, each displayed
task is evaluated with 50 trials per seed. The averages reported in Table~\ref{tab:main_results} are computed
from the corresponding seed-level success rates; suite-level and
task-level averages are reported as arithmetic means of the displayed
scores.

\begin{table}[!htbp]
\centering
\caption{Seed-level closed-loop success counts of SCULPT-VLA on LIBERO
and SimplerEnv-WidowX. LIBERO rows report successful trials out of 500
per seed, corresponding to 10 tasks with 50 trials each. SimplerEnv-WidowX
rows report successful trials out of 50 per seed for each displayed task.
Success Rate is mean $\pm$ sample SD over three seeds.}
\label{tab:simulation_seed_counts}
\scriptsize
\setlength{\tabcolsep}{4pt}
\renewcommand{\arraystretch}{1.12}
\resizebox{\textwidth}{!}{%
\begin{tabular}{lccc c}
\toprule
\textbf{Benchmark / Task} &
\textbf{Seed 1} &
\textbf{Seed 2} &
\textbf{Seed 3} &
\textbf{Success Rate (\%)} \\
\midrule
LIBERO-Spatial      & 496/500 & 494/500 & 497/500 & $99.1{\pm}0.3$ \\
LIBERO-Object       & 498/500 & 496/500 & 499/500 & $99.5{\pm}0.3$ \\
LIBERO-Goal         & 492/500 & 493/500 & 489/500 & $98.3{\pm}0.4$ \\
LIBERO-Long         & 485/500 & 482/500 & 487/500 & $96.9{\pm}0.5$ \\
\midrule
SimplerEnv-Spoon    & 45/50 & 45/50 & 46/50 & $90.7{\pm}1.2$ \\
SimplerEnv-Carrot   & 35/50 & 36/50 & 35/50 & $70.7{\pm}1.2$ \\
SimplerEnv-Stack    & 38/50 & 37/50 & 38/50 & $75.3{\pm}1.2$ \\
SimplerEnv-Eggplant & 49/50 & 49/50 & 48/50 & $97.3{\pm}1.2$ \\
\bottomrule
\end{tabular}}
\end{table}

For SimplerEnv-WidowX, each seed totals 167 successes in 200 trials.
The four-task aggregate is therefore $83.5\pm0.0$\%, although individual
task scores vary across seeds.

\subsection{Ablation Protocol and Variant Definitions}
\label{app:ablation_details}

\paragraph{Protocol.}
SimplerEnv-WidowX ablations report success as mean $\pm$ sample SD
over evaluation runs \citep{li2024evaluating}.
Unless changed by the intervention, variants share the Qwen3-VL
backbone, action data, horizon, optimizer, teacher targets,
FAST tokenizer \citep{pertsch2025fast}, flow head, and evaluation.
Each variant is trained in independent runs. FAST denotes the Stage-II
policy and continuous success the Stage-III policy. We report FAST
success for variants trained with the Stage-II FAST objective; a dash
denotes variants without that supervised Stage-II policy.
OLS/mOLS follows Appendix~\ref{app:open-loop}.
Table~\ref{tab:analysis_ablation_full} retains all variants and thresholds.
\begin{table}[!htbp]
  \centering
  \caption{Ablation results on SimplerEnv-WidowX Avg. We report
  closed-loop success under Stage-II FAST decoding and Stage-III
  continuous control, together with open-loop OLS and mOLS. Success rates
  are mean $\pm$ sample SD over evaluation runs; variants are trained
  independently.
  OLS and mOLS are defined in Appendix~\ref{app:open-loop}.
  Dashes indicate variants without a supervised Stage-II FAST policy.}
  \label{tab:analysis_ablation_full}
  \scriptsize
  \setlength{\tabcolsep}{5.0pt}
  \renewcommand{\arraystretch}{1.05}
  \resizebox{\textwidth}{!}{%
  \begin{tabular}{lccccccc}
    \toprule
    \textbf{Variant} &
    \multicolumn{2}{c}{\textbf{Success Rate (\%)}} &
    \multicolumn{4}{c}{\textbf{OLS} $\uparrow$} &
    \textbf{mOLS} $\uparrow$ \\
    \cmidrule(lr){2-3} \cmidrule(lr){4-7}
    & \textbf{FAST} & \textbf{Cont.}
    & \textbf{@0.1} & \textbf{@0.05} & \textbf{@0.03}
    & \textbf{@0.01} &  \\
    \midrule

    \textbf{Full SCULPT-VLA}
      & $74.7{\pm}1.0$ & $83.5{\pm}0.0$
      & 0.738 & 0.431 & 0.248
      & 0.057 & 0.369 \\

    shared factor FFN
      & $65.7{\pm}1.0$ & $76.8{\pm}1.0$
      & 0.697 & 0.379 & 0.201
      & 0.039 & 0.329 \\

    generic latent pool
      & $52.8{\pm}1.2$ & $63.9{\pm}1.1$
      & 0.632 & 0.301 & 0.132
      & 0.020 & 0.271 \\

    w/o subtask factor
      & $67.3{\pm}1.3$ & $78.3{\pm}1.3$
      & 0.713 & 0.397 & 0.216
      & 0.044 & 0.342 \\

    w/o spatial factor
      & $64.8{\pm}0.8$ & $72.2{\pm}0.8$
      & 0.680 & 0.356 & 0.183
      & 0.031 & 0.312 \\

    w/o dynamics factor
      & $61.6{\pm}1.0$ & $73.3{\pm}0.8$
      & 0.687 & 0.372 & 0.193
      & 0.036 & 0.322 \\

    w/o action-side latent slots $\mathbf{h}^{\mathrm{act}}$
      & $69.2{\pm}1.3$ & $77.0{\pm}1.0$
      & 0.706 & 0.386 & 0.207
      & 0.040 & 0.335 \\

    w/o Stage-II visibility restriction
      & $55.7{\pm}0.8$ & $71.3{\pm}1.0$
      & 0.671 & 0.342 & 0.168
      & 0.028 & 0.302 \\

    w/o Stage I
      & $60.0{\pm}1.0$ & $70.2{\pm}1.0$
      & 0.667 & 0.343 & 0.166
      & 0.027 & 0.301 \\

    Stage III only
      & --- & $61.0{\pm}1.0$
      & 0.621 & 0.287 & 0.121
      & 0.018 & 0.262 \\

    w/o FAST supervision
      & --- & $68.2{\pm}1.3$
      & 0.651 & 0.329 & 0.154
      & 0.025 & 0.290 \\

    w/o DINOv2 detail branch
      & $73.7{\pm}1.0$ & $81.0{\pm}1.0$
      & 0.724 & 0.414 & 0.229
      & 0.049 & 0.354 \\

    \bottomrule
  \end{tabular}}
\end{table}

\paragraph{Representation controls.}
The \emph{generic latent pool} removes explicit type identities,
factor-specific residual FFNs, and type-dependent visibility constraints.
Teacher targets remain assigned to the original $4$ subtask, $4$ dynamics,
and $8$ spatial positions. Token count, DINOv2, both action objectives,
and the three-stage schedule are unchanged.
This control tests explicit factor organization under the same supervision.

The \emph{shared factor FFN} replaces
$\Delta\mathrm{FFN}_{\mathrm{sub}}$,
$\Delta\mathrm{FFN}_{\mathrm{dyn}}$, and
$\Delta\mathrm{FFN}_{\mathrm{spa}}$ with one
$\Delta\mathrm{FFN}_{\mathrm{factor}}$.
Type labels, visibility, and the action branch
$\Delta\mathrm{FFN}_{\mathrm{act}}$ remain unchanged.

Each \emph{factor-removal} variant removes one factor's slots and
scaffold loss. The \emph{w/o action-side latent slots} variant removes
$\mathbf h^{\mathrm{act}}$ and decodes from the factor states.
The \emph{w/o Stage-II visibility restriction} variant permits
act-typed positions to read vision and language directly.

\paragraph{Training controls.}
\emph{w/o Stage I} skips alignment warm-up.
\emph{Stage III only} omits the first two stages while retaining the
SCULPT-VLA architecture and auxiliary losses, distinguishing it from
the direct Qwen3-VL-PI baseline.
\emph{w/o FAST supervision} removes the FAST objective while preserving staged training.

The \emph{w/o DINOv2 detail branch} variant removes the additional visual
features from the Stage-III head. Its Stage-II checkpoint comes from an
independent training run. Since the branch is introduced in Stage~III,
the FAST-score difference reflects variation between training runs.

\paragraph{Interpretation.}
The shared-FFN control isolates residual sharing under typed labels and
visibility, whereas the generic pool changes all three together.
These interventions test complementary design choices rather than
additive contributions. Stage-removal comparisons differ in total
optimization budget under the reported schedules.

\subsection{Real-World Experimental Setup}
\label{app:real_world_setup}

\paragraph{Hardware and control.}
We use a Universal Robots UR5e arm with a three-finger flexible gripper
on a $1.0 \times 0.5 \times 0.70~\mathrm{m}$ tabletop workspace. At
each policy query, the model outputs an action chunk of length $H=16$,
where each action contains a 6-DoF end-effector pose target and a
gripper command. The policy is queried at approximately
$7.8~\mathrm{Hz}$, and the resulting targets are executed by an RTDE
position controller running at $125~\mathrm{Hz}$. Visual observations
are provided by synchronized Intel RealSense D435i cameras, including a
wrist-mounted view for local interaction and a third-person view for
global scene context. Qualitative figures in this appendix are rendered
from the third-person view.

\paragraph{Task definitions and OOD settings.}
Table~\ref{tab:realworld_ood_details} summarizes the real-world tasks,
success criteria, and OOD perturbations. ID settings reuse the
demonstration setup, while OOD settings perturb object, layout, or visual
factors not seen in the corresponding demonstrations.

\begin{table}[!htbp]
\centering
\caption{Real-world task definitions and OOD perturbations.}
\label{tab:realworld_ood_details}
\scriptsize
\setlength{\tabcolsep}{4pt}
\renewcommand{\arraystretch}{1.15}
\resizebox{\textwidth}{!}{%
\begin{tabular}{p{2.5cm}p{4.1cm}p{4.0cm}p{4.3cm}}
\toprule
\textbf{Task} & \textbf{Success criterion} & \textbf{ID setting} & \textbf{OOD perturbations} \\
\midrule
Fruit Collecting &
All target fruits are picked and placed into the basket without dropping,
collision-induced failure, or premature termination. &
Training fruit instances and canonical basket layout. &
Unseen fruit instances, changed fruit order, and altered initial object
positions. \\

Paper Cup Stacking &
The target cup is lifted, aligned, and stably stacked on the base cup. &
Training cup colors/textures and nominal relative placement. &
Novel cup colors/textures, shifted cup positions, and altered approach
directions. \\

Toothbrush Insertion &
The toothbrush is grasped, transported, and inserted into the target
container with stable final placement. &
Training toothbrush/container pair and nominal insertion pose. &
Unseen toothbrush or container appearance, shifted container position,
and altered insertion orientation. \\

Cloth Folding &
The cloth is folded along the intended direction and remains approximately
aligned after release. &
Training cloth appearance and canonical tabletop placement. &
Changed cloth position/orientation, visual texture variation, and
perturbed initial spread. \\
\bottomrule
\end{tabular}}
\end{table}

\paragraph{Complete real-world results.}
Table~\ref{tab:real_world_full} supplements the main success table with
open-loop scores for every task and condition.
\begin{table}[!htbp]
\centering
\caption{Complete real-world open-loop and closed-loop results. Each task has ID and OOD columns for OLS@0.1, mOLS, and success rate (SC, \%). Open-loop scores use held-out expert trajectories. SC is mean $\pm$ sample SD across three independent evaluation batches of 30 physical rollouts per condition, using one checkpoint per method/task.}
\label{tab:real_world_full}
\footnotesize
\setlength{\tabcolsep}{2.8pt}
\renewcommand{\arraystretch}{1.1}
\begin{tabular}{llcccccc}
\toprule
& & \multicolumn{3}{c}{ID} & \multicolumn{3}{c}{OOD} \\
\cmidrule(lr){3-5}\cmidrule(lr){6-8}
Method & Task & OLS@0.1 & mOLS & SC & OLS@0.1 & mOLS & SC \\
\midrule
OpenVLA-OFT & Fruit Collecting & $0.648$ & $0.271$ & $50.0{\pm}3.3$ & $0.529$ & $0.203$ & $24.4{\pm}5.1$ \\
 & Paper Cup Stacking & $0.689$ & $0.298$ & $61.1{\pm}1.9$ & $0.596$ & $0.239$ & $36.7{\pm}6.7$ \\
 & Toothbrush Insertion & $0.629$ & $0.263$ & $44.4{\pm}1.9$ & $0.485$ & $0.184$ & $17.8{\pm}5.1$ \\
 & Cloth Folding & $0.556$ & $0.224$ & $25.6{\pm}1.9$ & $0.427$ & $0.129$ & $2.2{\pm}1.9$ \\
$\pi_0$ & Fruit Collecting & $0.704$ & $0.319$ & $65.6{\pm}1.9$ & $0.637$ & $0.276$ & $45.6{\pm}6.9$ \\
 & Paper Cup Stacking & $0.771$ & $0.353$ & $75.6{\pm}1.9$ & $0.668$ & $0.302$ & $56.7{\pm}3.3$ \\
 & Toothbrush Insertion & $0.724$ & $0.337$ & $67.8{\pm}3.8$ & $0.619$ & $0.282$ & $44.4{\pm}5.1$ \\
 & Cloth Folding & $0.641$ & $0.273$ & $35.6{\pm}1.9$ & $0.536$ & $0.185$ & $14.4{\pm}3.8$ \\
$\pi_{0.5}$ & Fruit Collecting & $0.741$ & $0.345$ & $74.4{\pm}1.9$ & $0.653$ & $0.292$ & $52.2{\pm}5.1$ \\
 & Paper Cup Stacking & $0.796$ & $0.362$ & $81.1{\pm}1.9$ & $0.705$ & $0.323$ & $60.0{\pm}6.7$ \\
 & Toothbrush Insertion & $0.748$ & $0.351$ & $73.3{\pm}3.3$ & $0.646$ & $0.294$ & $47.8{\pm}3.8$ \\
 & Cloth Folding & $0.665$ & $0.288$ & $41.1{\pm}1.9$ & $0.574$ & $0.229$ & $22.2{\pm}5.1$ \\
SCULPT-VLA & Fruit Collecting & $0.758$ & $0.353$ & $80.0{\pm}3.3$ & $0.681$ & $0.304$ & $60.0{\pm}3.3$ \\
 & Paper Cup Stacking & $0.791$ & $0.366$ & $85.6{\pm}1.9$ & $0.732$ & $0.336$ & $70.0{\pm}6.7$ \\
 & Toothbrush Insertion & $0.762$ & $0.356$ & $84.4{\pm}5.1$ & $0.688$ & $0.308$ & $64.4{\pm}5.1$ \\
 & Cloth Folding & $0.691$ & $0.299$ & $56.7{\pm}3.3$ & $0.603$ & $0.227$ & $37.8{\pm}5.1$ \\
\bottomrule
\end{tabular}
\end{table}

\subsection{Statistical Reporting for Real-World Evaluation}
\label{app:realworld_stats}

\paragraph{Open-loop validation data.}
For real-world open-loop evaluation, we use held-out expert
trajectories collected separately for ID and OOD settings. Training uses
only the ID demonstrations reported in the main text. The ID validation
trajectories follow the training distribution, while the OOD validation
trajectories follow the same perturbation families used for the
closed-loop OOD evaluation in Table~\ref{tab:realworld_ood_details}.
The validation trajectories are reserved for OLS/mOLS computation and
excluded from training, scaffold construction, and model selection.
During open-loop evaluation, the policy receives only the observation
and instruction at each timestep; expert action chunks are used solely
as references for computing OLS/mOLS following
Appendix~\ref{app:open-loop}.

\paragraph{Closed-loop statistical reporting.}
Each method uses one trained checkpoint per task. For each ID/OOD
condition, success (SC) is measured in three independent evaluation
batches of $30$ physical rollouts, totaling $90$ rollouts per condition.
Mean $\pm$ sample SD reports variation across these three evaluation
batches.
Table~\ref{tab:realworld_raw_counts} gives the counts.

\begin{table}[!htbp]
\centering
\caption{Real-world SC counts across evaluation batches. Each entry
reports successful rollouts out of $30$ for Batch 1/Batch 2/Batch 3,
using one trained checkpoint per method and task.}
\label{tab:realworld_raw_counts}
\scriptsize
\setlength{\tabcolsep}{3pt}
\renewcommand{\arraystretch}{1.1}
\resizebox{\textwidth}{!}{%
\begin{tabular}{lcccccccc}
\toprule
\textbf{Method} &
\textbf{Fruit ID} & \textbf{Fruit OOD} &
\textbf{Cup ID} & \textbf{Cup OOD} &
\textbf{Toothbrush ID} & \textbf{Toothbrush OOD} &
\textbf{Cloth ID} & \textbf{Cloth OOD} \\
\midrule
OpenVLA-OFT &
15/14/16 & 7/9/6 & 19/18/18 & 11/9/13 &
13/14/13 & 5/7/4 & 8/8/7 & 1/1/0 \\
$\pi_0$ &
20/19/20 & 13/16/12 & 23/22/23 & 17/16/18 &
21/19/21 & 13/15/12 & 11/11/10 & 5/5/3 \\
$\pi_{0.5}$ &
22/22/23 & 16/14/17 & 24/25/24 & 18/20/16 &
22/23/21 & 15/15/13 & 13/12/12 & 8/7/5 \\
SCULPT-VLA &
24/25/23 & 18/17/19 & 25/26/26 & 21/23/19 &
25/24/27 & 19/21/18 & 18/17/16 & 13/11/10 \\
\bottomrule
\end{tabular}}
\end{table}

\paragraph{Variation across OOD tasks.}
The OOD gains over $\pi_{0.5}$ are largest on Toothbrush Insertion
($+16.7$\,pp) and Cloth Folding ($+15.6$\,pp), followed by Paper Cup
Stacking ($+10.0$\,pp) and Fruit Collecting ($+7.8$\,pp).
These tasks differ in both interaction demands and perturbation
composition (Table~\ref{tab:realworld_ood_details}).
The task-level results characterize performance under these combined
perturbations.

\section{Functional Analysis of Learned Structured Control States}
\label{app:internalization_analysis}
The following analyses distinguish group geometry, decodable content,
attention allocation, and functional policy dependence.
They supplement Section~\ref{sec:exp-qual} of the main paper.

\subsection{Analysis Setting}
\label{app:analysis_setting}

The stage-wise geometry and group-level attention analyses use the
LIBERO 40-task video collection \citep{liu2023libero}.
We decode each video frame by frame and extract final-layer hidden
states and attention. Task-specific probes, trajectory analysis,
and the real-world intervention use the settings described below.
Tokens are partitioned into six groups:
visual (\texttt{vis}), instruction (\texttt{lang}), subtask
(\texttt{sub}), dynamics (\texttt{dyn}), spatial (\texttt{spa}),
and action-side (\texttt{act}).
The action group includes fixed readout slots
$\mathbf h^{\mathrm{act}}$ and, in discrete-policy forwards, FAST positions.
Stage-III continuous forwards contain no FAST positions.

For geometry visualization, we average the final hidden states within
each group to obtain a frame-level representation:
\begin{equation}
\mathbf{h}_{g}^{(t)}
=
\frac{1}{|S_g^{(t)}|}
\sum_{i\in S_g^{(t)}} \mathbf{h}_{i}^{(t)},
\end{equation}
where $S_g^{(t)}$ is the set of tokens belonging to group $g$ in
frame $t$. Stage~I, Stage~II, and Stage~III are embedded into a
shared PCA+UMAP space so that their low-dimensional geometries are
directly comparable. Silhouette scores are computed in the original
hidden space, avoiding distortion from the nonlinear 2D UMAP projection.

For routing analysis, we use the final-layer attention matrix after
head averaging. Given a query group $q$ and a key group $k$, we compute
the query-normalized attention mass routed from $q$ to $k$:
\begin{equation}
A(q,k)
=
\frac{1}{|Q_q|}
\sum_{i\in Q_q}\sum_{j\in K_k}\mathrm{Attn}_{ij},
\end{equation}
where $Q_q$ and $K_k$ denote the sets of query and key positions
belonging to groups $q$ and $k$, respectively. Since attention weights
are normalized over key positions for each query token, $A(q,k)$
measures the fraction of attention mass from an average query in group
$q$ assigned to key group $k$. These values are averaged over all frames
to form the stage-level Sankey diagrams.

\subsection{Stage-Wise Latent Geometry}
\label{app:stagewise_geometry}
\begin{figure}[htbp]
\centering
\begin{subfigure}{0.49\linewidth}
\includegraphics[width=\linewidth]{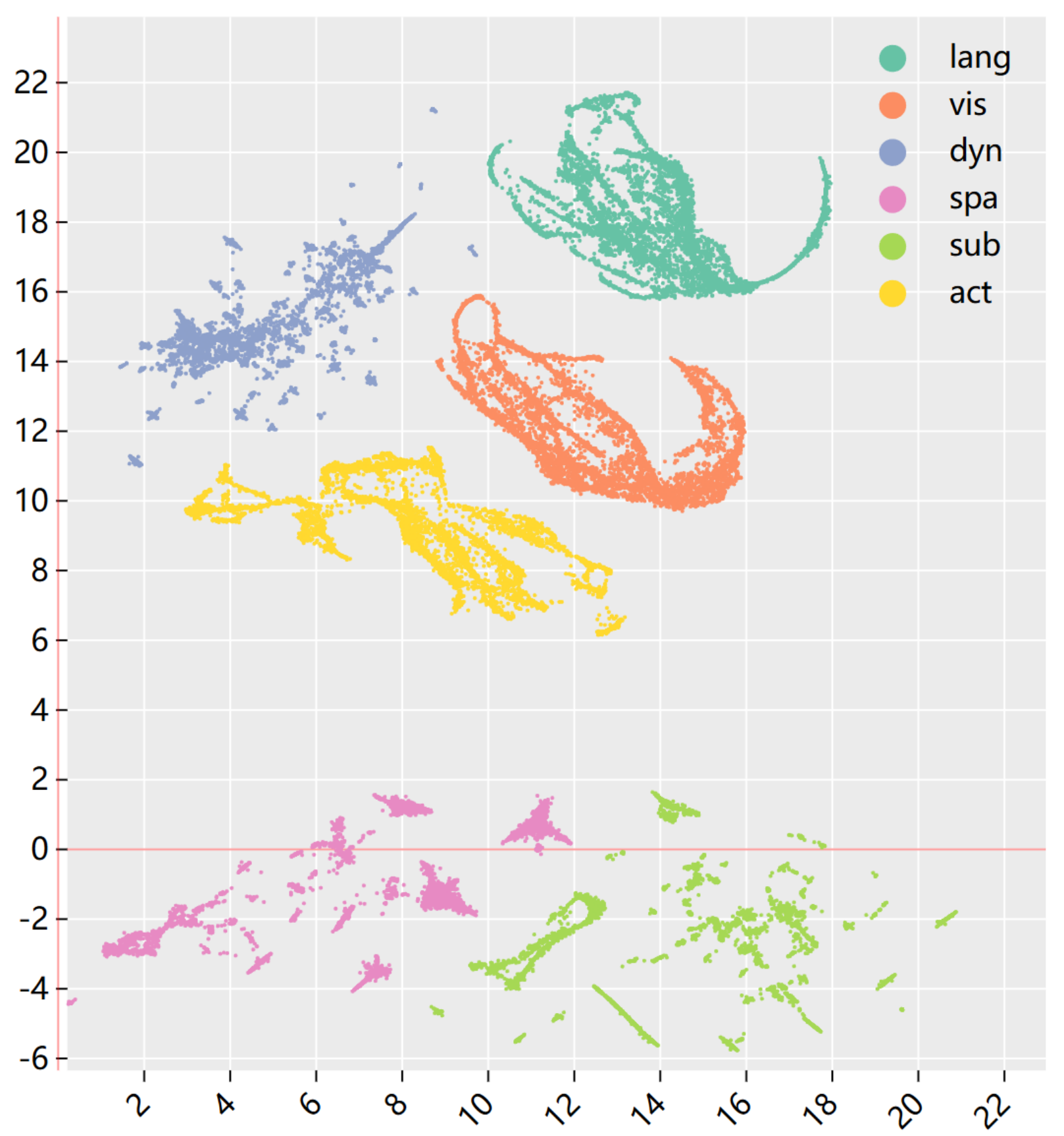}
\caption{Stage-III group representations.}\label{fig:stage3_umap}
\end{subfigure}\hfill
\begin{subfigure}{0.49\linewidth}
\includegraphics[width=\linewidth]{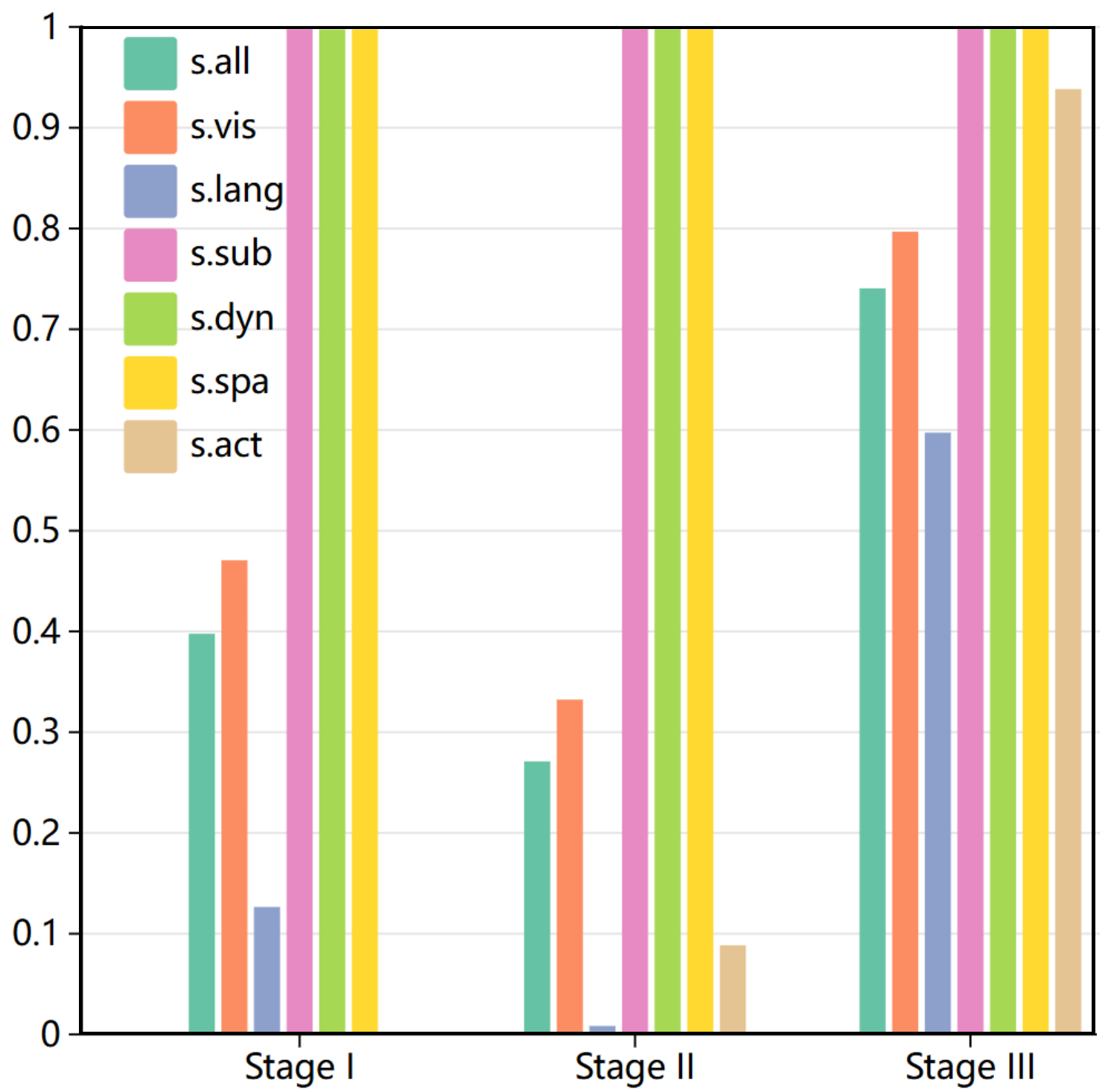}
\caption{Per-group silhouette scores.}\label{fig:silhouette_main}
\end{subfigure}
\caption{Group-level representation diagnostics. UMAP uses frame-level
group means. Silhouette scores are calculated in the original hidden
space.}
\end{figure}

\begin{figure}[!htbp]
  \centering
  \includegraphics[width=0.8\textwidth]{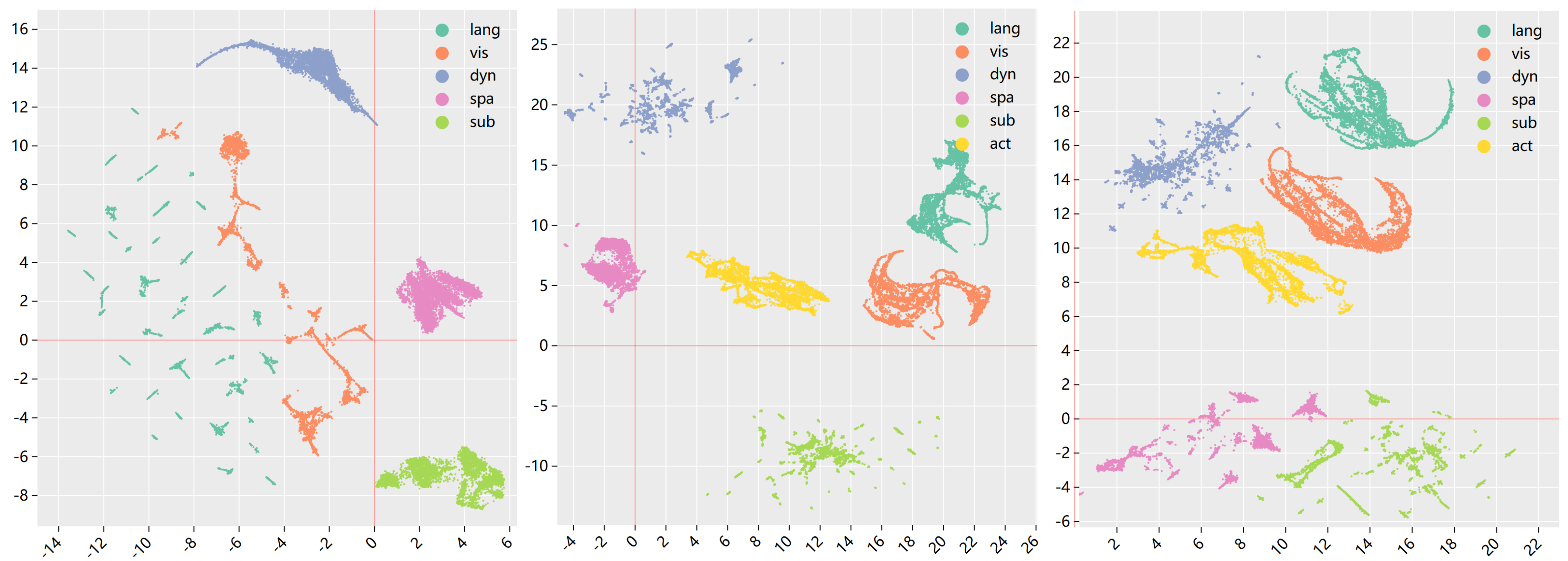}
  \caption{All-stage group-level UMAP in a shared PCA+UMAP space.
  Points are frame-level means for the active token groups;
  the action group appears from Stage~II onward. All stages share
  the same fitted projection.}
  \label{fig:allstage_umap_appendix}
\end{figure}

\begin{figure}[!htbp]
  \centering
  \includegraphics[width=0.8\textwidth]{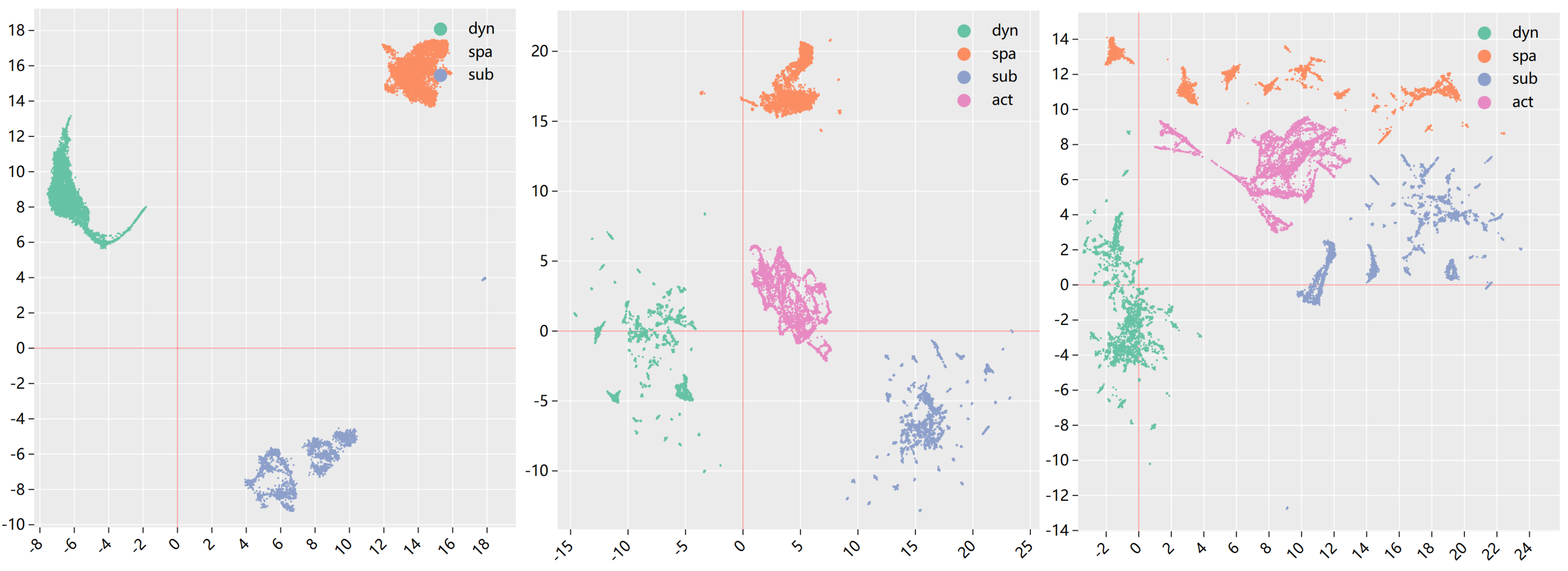}
  \caption{Factor- and action-group UMAP across training stages.
  The three factor groups occupy distinct regions from Stage~I onward;
  the action group is included in Stages~II and III.}
  \label{fig:slotonly_umap_appendix}
\end{figure}

Figs.~\ref{fig:allstage_umap_appendix}--\ref{fig:slotonly_umap_appendix}
extend the Stage~III overview into a complete
three-stage view, separating an all-group UMAP from a factor/action-group UMAP.

The factor/action-group projection separates the three factor groups at Stage~I.
The all-group projection changes most at Stage~II and shows more distinct
groups at Stage~III. These patterns describe the projected geometry of
the frame-level group representations, whose separation is encouraged
by factor-specific embeddings and teacher supervision.

Silhouette scores in the original hidden space complement the
projected geometry (Fig.~\ref{fig:silhouette_main}).
The factor groups remain distinguishable across stages.
Visual and language groups show lower separation in Stage~II and
recover in Stage~III, while action-group separation increases from
Stage~II to Stage~III.

\subsection{Routing Dynamics Across Stages}
\label{app:routing_dynamics}
\begin{figure}[htbp]
\centering
\includegraphics[width=\linewidth]{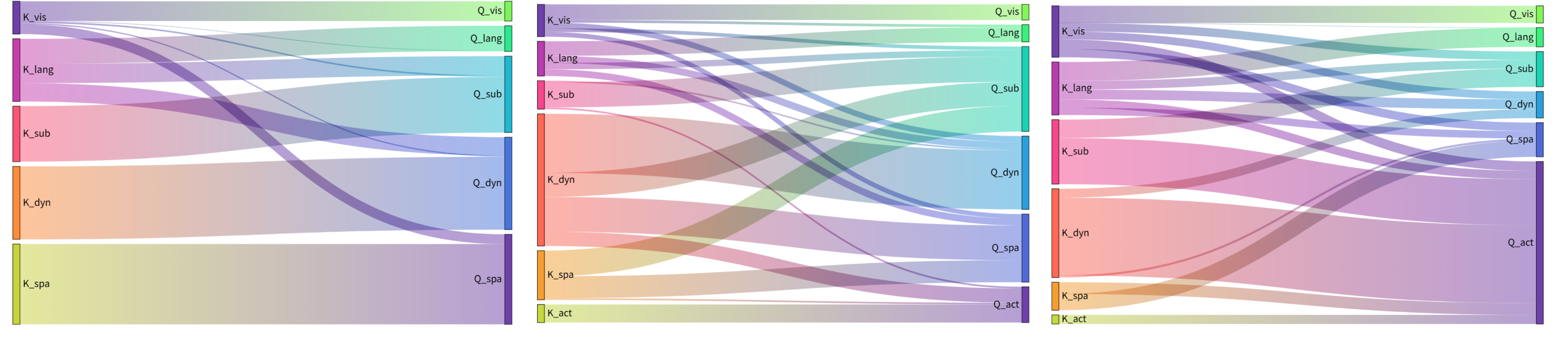}
\caption{Stage-wise final-layer attention mass. Panels show Stages I,
II, and III. Left nodes are key groups and right nodes are query groups.
Band widths represent attention mass between groups.}
\label{fig:sankey_main}
\end{figure}

The Sankey diagrams summarize final-layer attention mass between token
groups. Stage~I blocks cross-factor attention, so its within-group
pattern is partly determined by the mask. Stage~II permits cross-factor
attention but blocks direct vision/language access by action tokens.
Only Stage~III permits both routes. Attention describes information
allocation under each mask; factor-wise intervention separately tests
the policy's functional use of the resulting states.

\subsubsection{Effect of Stage-II Perceptual Access}
\label{app:stage3_shortcut}

Stage~III restores direct vision/language access for action tokens.
To examine the effect of introducing this access earlier,
we compare the standard schedule with a variant that uses
$\mathbf M^{(3)}$ in both Stages~II and III.
On LIBERO-Spatial, success is 98.7\% for this variant and 99.1\%
for the standard schedule (Table~\ref{tab:shortcut_mask_control}).

\begin{table}[!htbp]
\centering
\caption{Stage-II perceptual-access diagnostic on LIBERO-Spatial.
Success is the three-seed mean under the main evaluation protocol.
The w/o Stage-II visibility restriction variant uses $\mathbf M^{(3)}$ in both Stages~II
and III, allowing act-typed positions to read vision and language
throughout action grounding.}
\label{tab:shortcut_mask_control}
\scriptsize
\setlength{\tabcolsep}{7pt}
\renewcommand{\arraystretch}{1.1}
\begin{tabular}{lccc}
\toprule
\textbf{Variant} & \textbf{Stage-II Mask} & \textbf{Stage-III Mask} &
\textbf{LIBERO-Spatial Success (\%)} \\
\midrule
Standard SCULPT-VLA & $\mathbf{M}^{(2)}$ & $\mathbf{M}^{(3)}$ & $99.1$ \\
w/o Stage-II visibility restriction & $\mathbf{M}^{(3)}$ & $\mathbf{M}^{(3)}$ & $98.7$ \\
\bottomrule
\end{tabular}
\end{table}

The effect is smaller than on SimplerEnv-WidowX, indicating that
the benefit of restricted Stage-II access depends on the task.

\subsection{Factor Specialization and Functional Use}
\label{app:latent_analysis}

Spatial and dynamics probes characterize decodable content, while
trajectory-level subtask geometry and slot-to-visual attention describe
factor specialization. Inference-time replacement tests how final factor
states contribute to the Stage-III continuous policy.

\subsubsection{Spatial and Dynamics Probes}
\label{app:spa_dyn_probe}

\paragraph{Probe design.}
The spatial probe reconstructs a dense current-scene depth map. The dynamics probe
predicts a depth-residual map to characterize task-relevant state
transitions. Both probes are lightweight readouts trained separately
with the policy frozen, and provide qualitative visualizations of the
learned representations.

\begin{figure*}[!htbp]
  \centering
  \includegraphics[width=\textwidth]{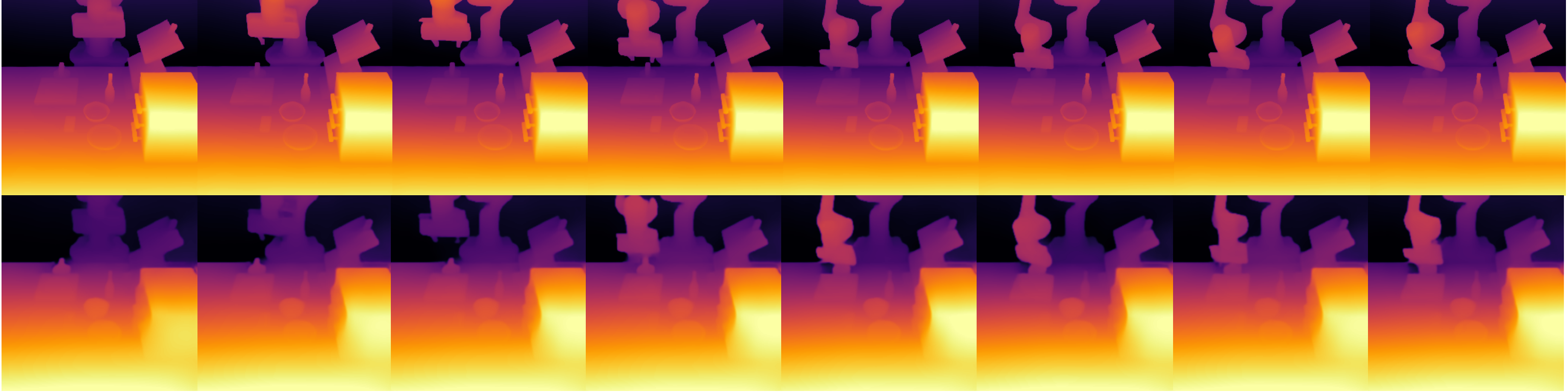}
  \caption{Spatial-slot probe against the depth target. Top: target
  depth map of the current observation; bottom: depth map
  reconstructed from spatial slots. The qualitative agreement in
  overall geometry is consistent with spatial slots preserving dense
  scene structure and spatial priors.}
  \label{fig:probe_spatial_appendix}
\end{figure*}

\paragraph{Spatial-slot probe.}
A lightweight probe maps spatial-slot states to a depth prediction
$\hat D_t^{\mathrm{spa}}$.
Figure~\ref{fig:probe_spatial_appendix} compares it with the current-depth
target; the reconstruction preserves the tabletop, manipulator silhouette,
and coarse object layout.

\paragraph{Dynamics-slot probe.}
To probe transition information, we construct a target from
local depth changes between frames.
Given two neighboring frames with monocular depth estimates $D_t$ and
$D_{t+\Delta}$, we first perform affine alignment to mitigate global
scale drift. We use $\Delta$ corresponding to approximately one second
in the processed videos.
\begin{equation}
D'_{t+\Delta} = a D_{t+\Delta} + b,
\end{equation}
where $(a,b)$ are obtained by least-squares fitting so that
$D'_{t+\Delta}$ best matches $D_t$. We then compute the absolute
residual
\begin{equation}
R_t = \left| D_t - D'_{t+\Delta} \right|
\end{equation}
To suppress high-frequency noise, we apply Gaussian smoothing to
$R_t$ with $\sigma=1.0$, threshold the smoothed residual at its
$85$th percentile, apply binary opening with a $3\times3$ structuring
element, and remove connected components smaller than $0.5\%$ of the
image area. The processed residual provides a sparse target for the dynamics probe,
emphasizing salient local depth changes. Its predicted visualization is
$\widehat{\Delta D}_{t}^{\mathrm{dyn}}$.

\paragraph{Probe observations.}
The spatial reconstruction shows overall scene geometry, whereas
the dynamics probe responds around changing interaction regions
(Fig.~\ref{fig:probe_main}).
These separately trained probes characterize decodable geometry and
transition information under their respective targets.

\subsubsection{Subtask-Slot PCA Analysis}
\label{app:subtask_pca}

\paragraph{Analysis method.}
We use trajectory-level PCA to examine whether subtask-slot
representations track execution phases in latent space.

\paragraph{Data construction.}
We perform this analysis on the \textit{turn\_on\_the\_stove} task
using $100$ trajectories. Each trajectory is divided post hoc into five
execution phases: \textit{reach},
\textit{align}, \textit{grasp}, \textit{turn}, and \textit{finish}.
These post-hoc labels are used only to color the extracted representations
and are independent of the object-switch annotations used for training.
For each frame, we extract the hidden states of the subtask-slot
tokens and average them over the slot dimension to obtain a
frame-level subtask representation:
\begin{equation}
\bar{\mathbf{h}}_{t}^{\mathrm{sub}}
=
\frac{1}{N_{\mathrm{sub}}}
\sum_{i=1}^{N_{\mathrm{sub}}}
\mathbf{h}_{t,i}^{\mathrm{sub}},
\end{equation}
where $N_{\mathrm{sub}}$ is the number of subtask-slot tokens
($N_{\mathrm{sub}}{=}4$). Frame-level representations from all
trajectories are aggregated and projected into PCA space for
visualization.

\begin{figure*}[!htbp]
  \centering
  \includegraphics[width=\textwidth]{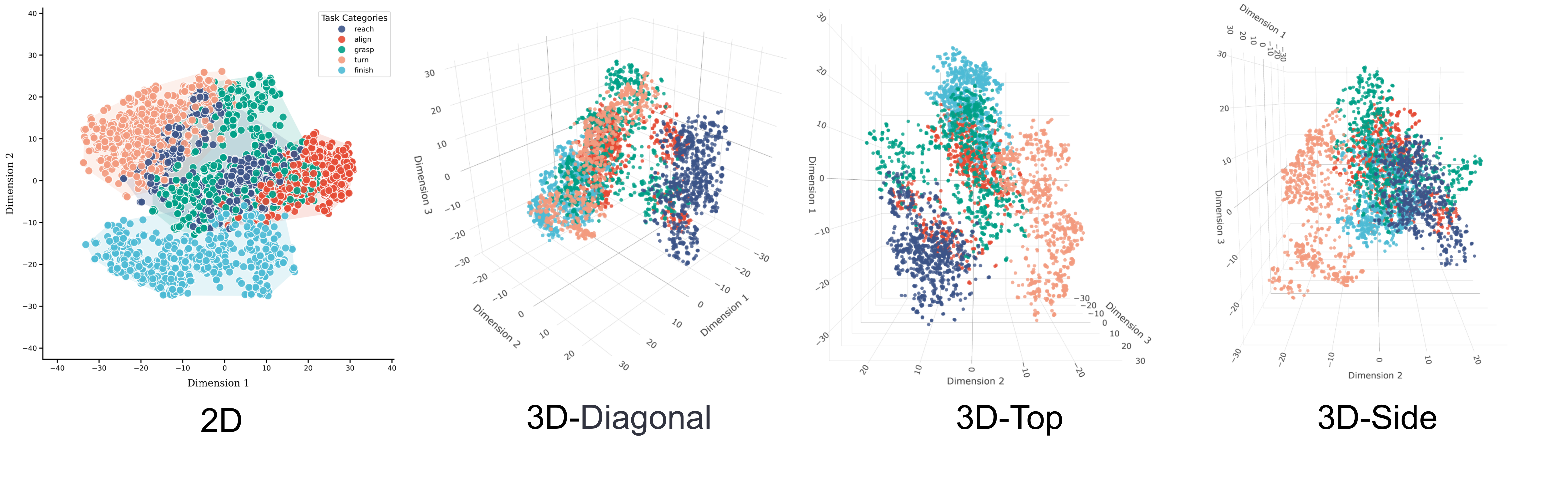}
  \caption{PCA of frame-level subtask-slot means from $100$
trajectories of \textit{turn\_on\_the\_stove}.
Colors indicate five post-hoc, analysis-only execution phases.
The 2D and 3D views show phase-associated geometry,
including connected neighboring regions and a compact terminal region.}
\label{fig:subtask_pca_appendix}
\end{figure*}

\paragraph{Phase-associated geometry.}
The projected representations vary with the annotated execution phases.
Adjacent phases occupy connected regions, while \textit{finish} is
comparatively compact.
The 3D view separates some regions that overlap in 2D
(Fig.~\ref{fig:subtask_pca_appendix}).
These patterns associate subtask-slot geometry with execution progress.

\subsubsection{Attention-Based Slot Specialization}
\label{app:attention_vis}

\begin{figure}[!htbp]
  \centering
  \includegraphics[width=0.5\textwidth]{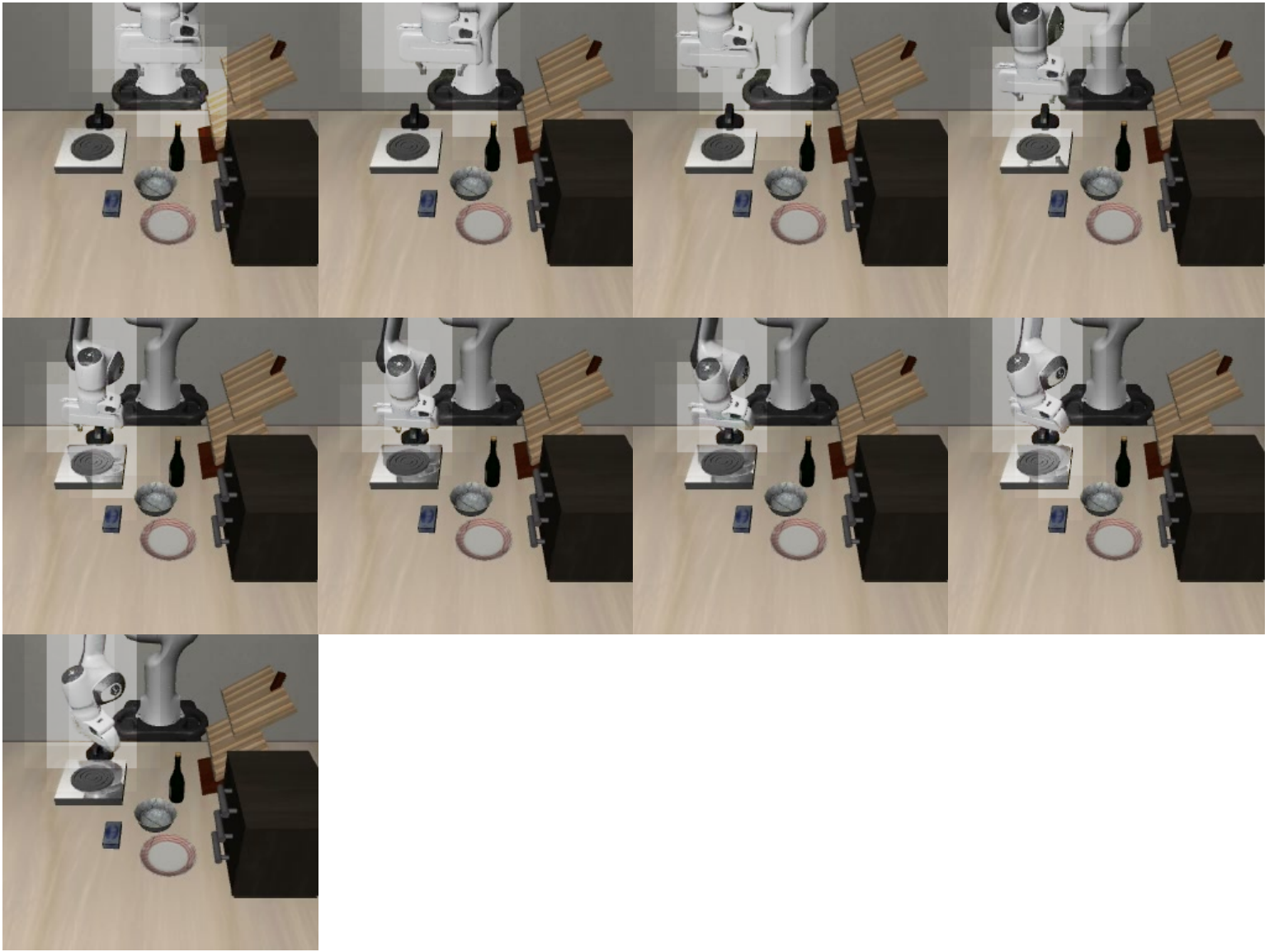}
  \caption{Dynamics-branch slot-to-visual attention on
  \textit{turn\_on\_the\_stove}. The high-response region shifts with
  the evolving end-effector--knob interaction, consistent with
  sensitivity to motion-dependent state change.}
  \label{fig:attn_dyn}
\end{figure}

\begin{figure}[!htbp]
  \centering
  \includegraphics[width=0.5\textwidth]{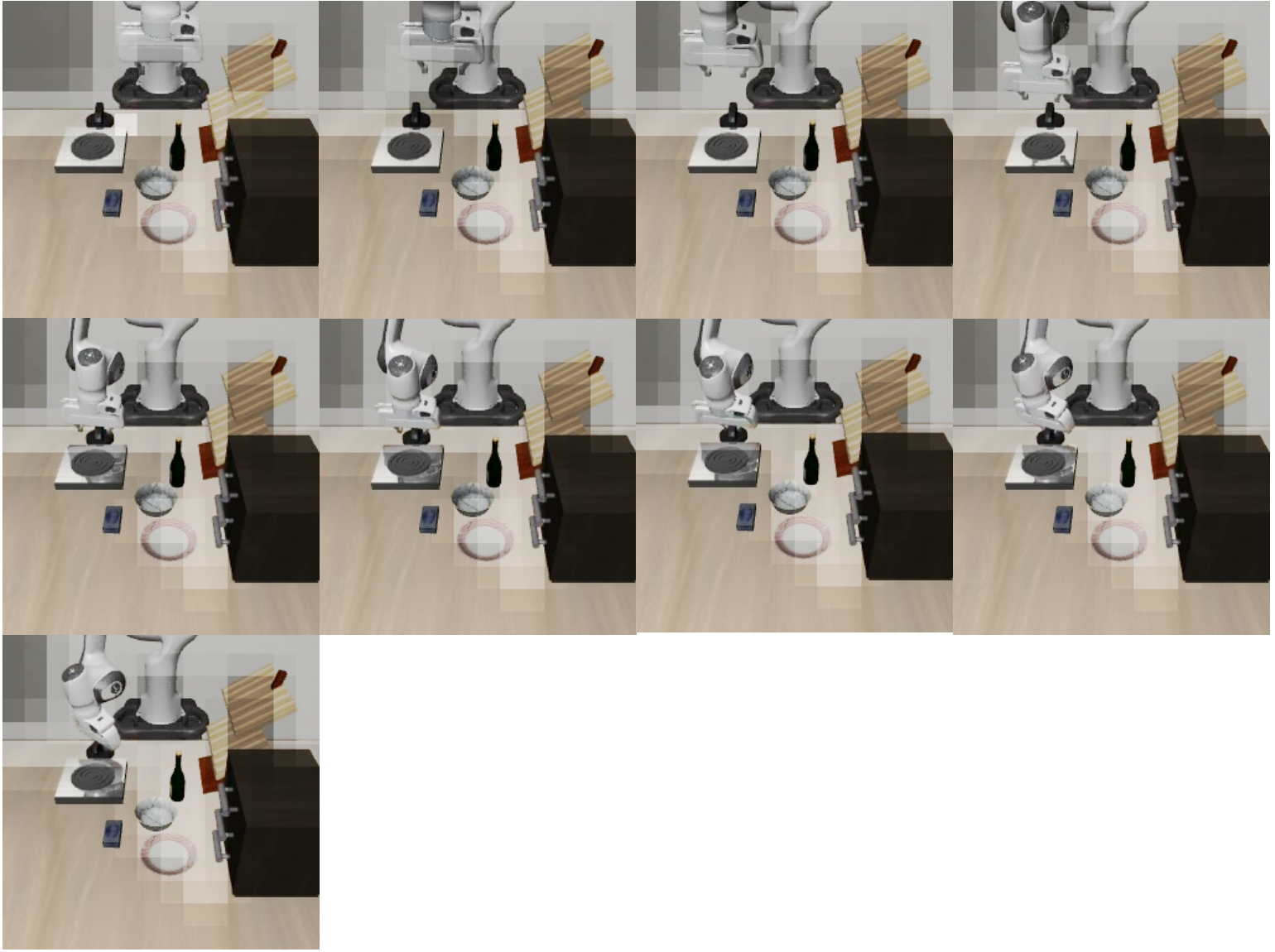}
  \caption{Spatial-branch slot-to-visual attention on
  \textit{turn\_on\_the\_stove}. Attention remains concentrated
  around the target knob and its geometric neighborhood, consistent
  with a role in target localization and local geometric grounding.}
  \label{fig:attn_spatial}
\end{figure}

\begin{figure}[!htbp]
  \centering
  \includegraphics[width=0.5\textwidth]{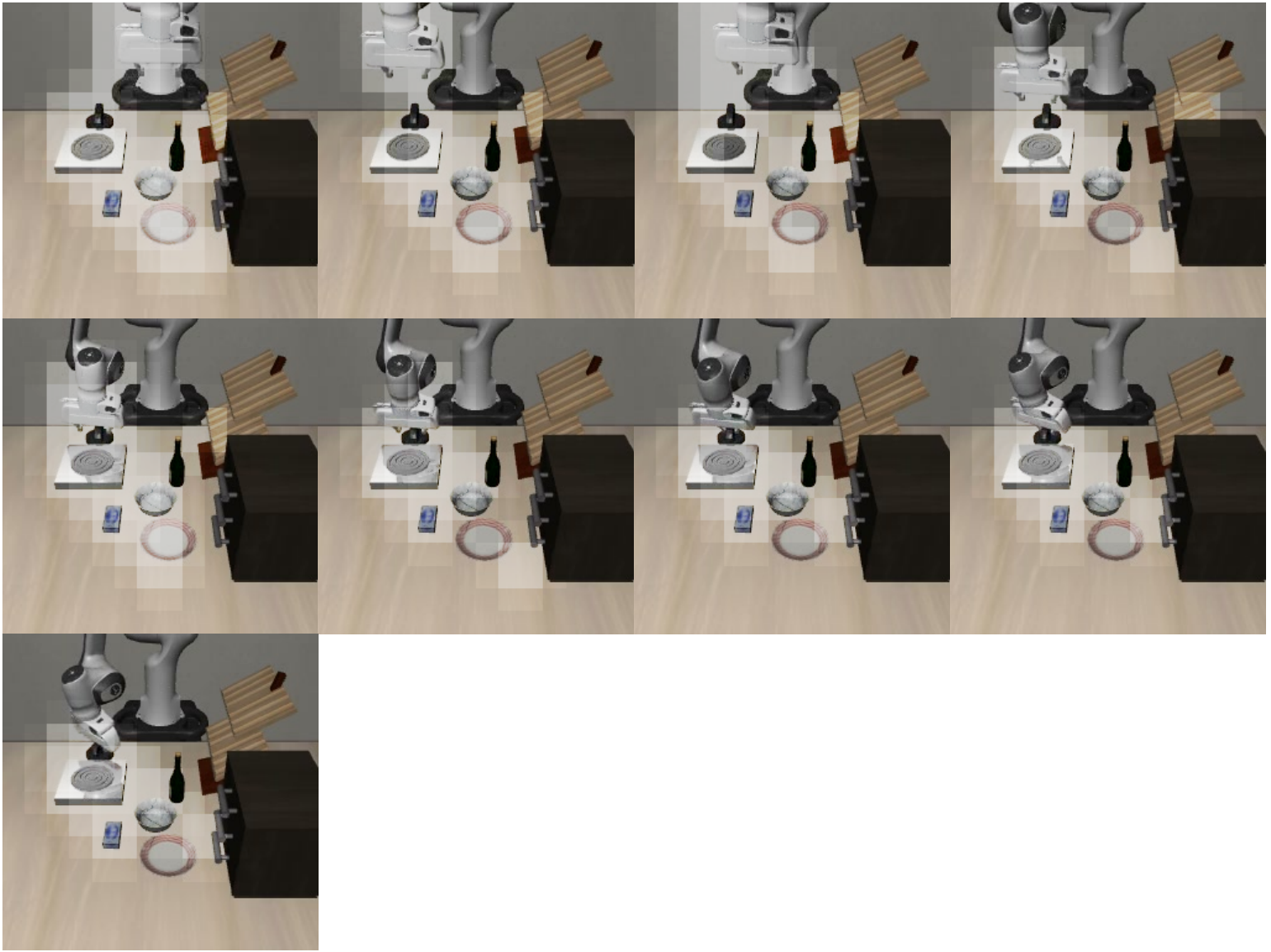}
  \caption{Subtask-branch slot-to-visual attention on
  \textit{turn\_on\_the\_stove}. Attention is broader and temporally
  more stable, covering the stove and surrounding semantic context,
  consistent with stage-level task progression modeling.}
  \label{fig:attn_sub}
\end{figure}

We visualize final-layer slot-to-visual attention
(Figs.~\ref{fig:attn_dyn}--\ref{fig:attn_sub}) to examine factor
specialization. For each branch
$b \in \{\mathrm{sub}, \mathrm{dyn}, \mathrm{spa}\}$, the
branch-to-visual attention response is
\[
\bar{\mathbf{a}}^{(b)}
=
\frac{1}{|\mathcal{H}|\,|\mathcal{S}_b|}
\sum_{h \in \mathcal{H}}
\sum_{s \in \mathcal{S}_b}
\mathbf{A}^{(L,h)}_{s,\mathcal{V}},
\]
where $\mathcal{S}_b$ and $\mathcal{V}$ are the branch slot and
visual token sets, and $\mathbf{A}^{(L,h)}$ is the final-layer
attention matrix. The $7\times7$ visual token response map is
min-max normalized, threshold-masked, and bilinearly upsampled to
$14\times14$ for display. The overlays retain the coarse spatial
resolution of the visual token grid.

\subsubsection{Factor-Wise Functional Intervention}
\label{app:slot_intervention}

We test the functional role of structured control factors in the
Stage-III continuous policy on the real-world Fruit Collecting task.
After the backbone forward pass, we replace the final hidden states of
a selected factor group with its training-set mean before continuous
action prediction. The mean is computed separately for each query
position and factor type.

Fruit Collecting requires repeated pick-and-place cycles and provides
a setting for comparing group-specific interventions.
We replace each factor separately and all factors jointly, retaining
action readout states and DINOv2 features.
Table~\ref{tab:slot_replacement} reports successful trials and
variation across evaluation batches using the same trained checkpoint.

\begin{table}[!htbp]
\centering
\caption{Factor-wise functional intervention on the real-world Fruit
Collecting task. Each evaluation batch reports successful rollouts out
of $30$ under the same Stage-III checkpoint. SC is mean $\pm$ sample SD
over three independent evaluation batches per condition.}
\label{tab:slot_replacement}
\scriptsize
\setlength{\tabcolsep}{5pt}
\renewcommand{\arraystretch}{1.12}
\resizebox{0.86\textwidth}{!}{%
\begin{tabular}{lcccccccc}
\toprule
\multirow{2}{*}{\textbf{Variant}} &
\multicolumn{4}{c}{\textbf{Fruit Collecting ID}} &
\multicolumn{4}{c}{\textbf{Fruit Collecting OOD}} \\
\cmidrule(lr){2-5} \cmidrule(lr){6-9}
& \textbf{Batch 1} & \textbf{Batch 2} & \textbf{Batch 3} & \textbf{SC (\%)}
& \textbf{Batch 1} & \textbf{Batch 2} & \textbf{Batch 3} & \textbf{SC (\%)} \\
\midrule

Full SCULPT-VLA
& 24/30 & 25/30 & 23/30 & $80.0{\pm}3.3$
& 18/30 & 17/30 & 19/30 & $60.0{\pm}3.3$ \\

Replace subtask factor
& 16/30 & 15/30 & 18/30 & $54.4{\pm}5.1$
& 8/30 & 12/30 & 6/30 & $28.9{\pm}10.2$ \\

Replace spatial factor
& 22/30 & 23/30 & 21/30 & $73.3{\pm}3.3$
& 15/30 & 13/30 & 17/30 & $50.0{\pm}6.7$ \\

Replace dynamics factor
& 20/30 & 21/30 & 18/30 & $65.6{\pm}5.1$
& 12/30 & 15/30 & 11/30 & $42.2{\pm}6.9$ \\

Replace all factors
& 5/30 & 3/30 & 6/30 & $15.6{\pm}5.1$
& 2/30 & 0/30 & 3/30 & $5.6{\pm}5.1$ \\

\bottomrule
\end{tabular}}
\end{table}

Task-progression replacement causes the largest drop among single-group
interventions, followed by dynamics and spatial replacement.
Replacing all three groups leaves low success despite retaining
action readout states and DINOv2 features, providing behavioral evidence
that the learned structured control state remains functionally involved
after direct perceptual access is restored.

Mean replacement alters both factor content and the conditioning
distribution, so the measured changes reflect the full intervention.

\FloatBarrier
\section{Additional Rollouts}
\label{app:qualitative_rollouts}
\subsection{Additional Simulation Visualizations}
\label{app:sim_vis}

\begin{figure}[H]
  \centering
  \includegraphics[width=0.5\textwidth]{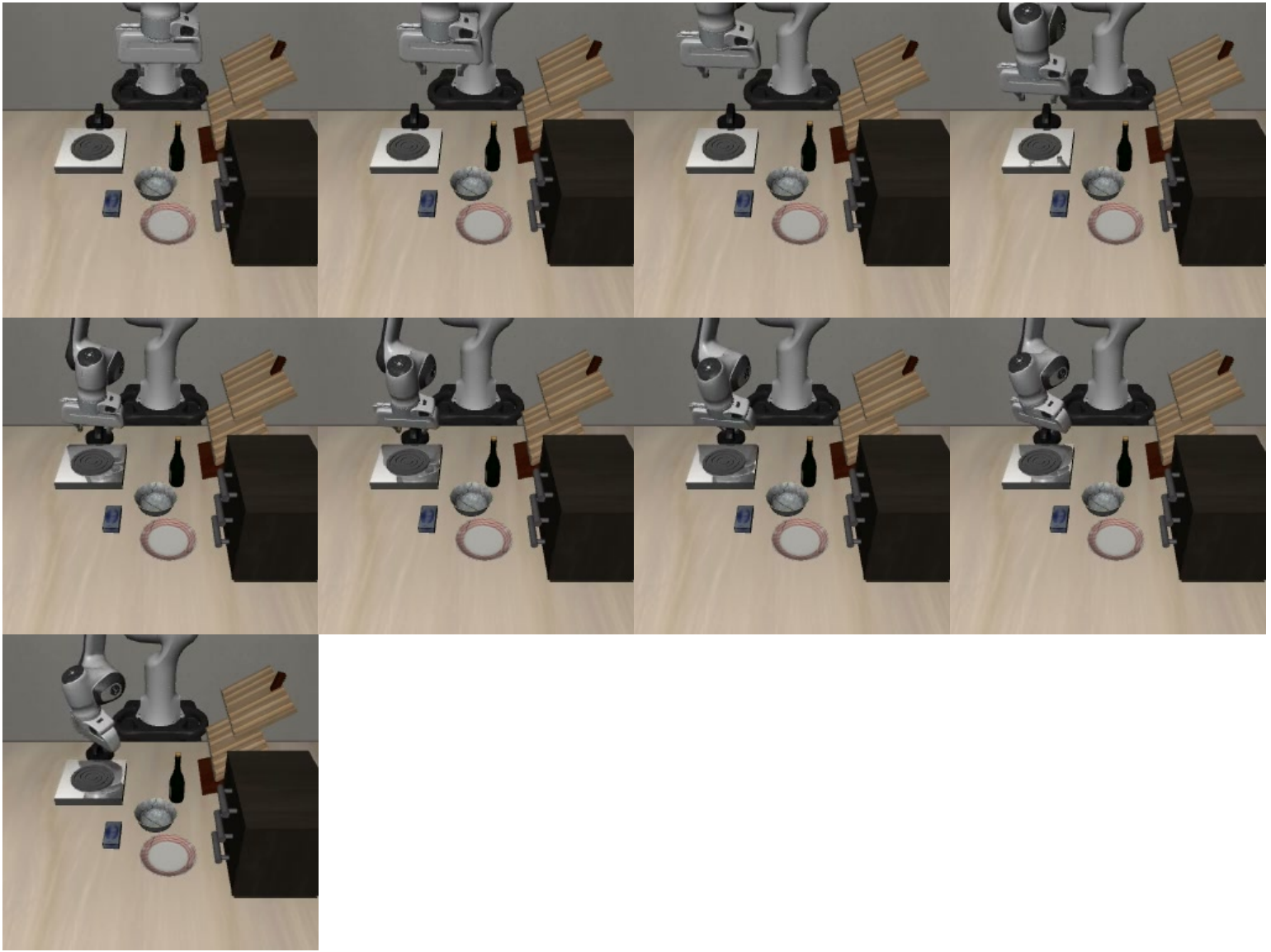}
  \caption{LIBERO \textit{turn\_on\_the\_stove}: approach,
  end-effector alignment with the knob, and contact-sensitive knob turning.}
  \label{fig:libero_stove_rollout}
\end{figure}

\begin{figure}[H]
  \centering
  \includegraphics[width=0.5\textwidth]{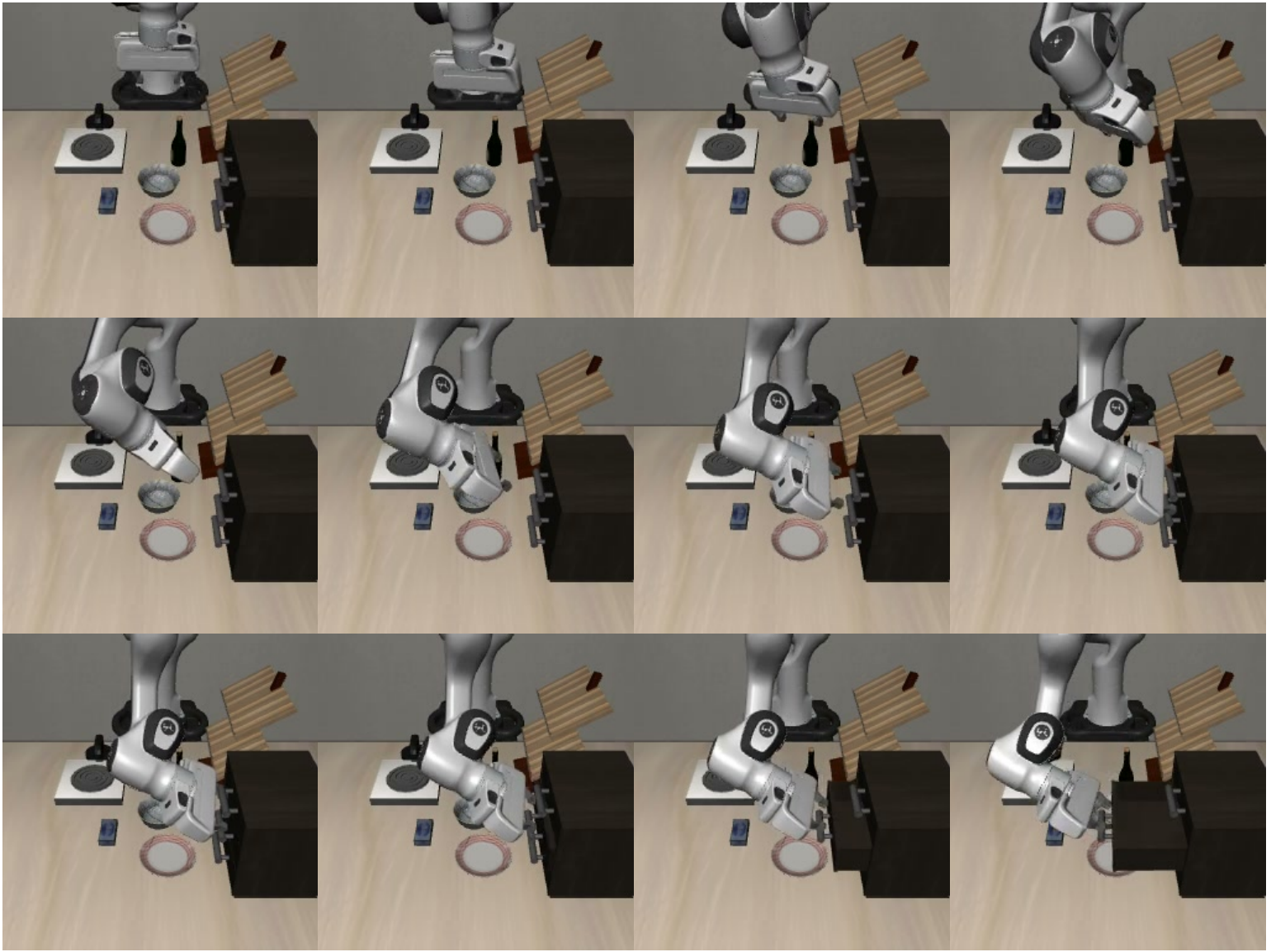}
  \caption{LIBERO
  \textit{open\_the\_middle\_drawer\_of\_the\_cabinet}: handle
  approach and sustained drawer pulling while maintaining contact
  with the handle.}
  \label{fig:libero_drawer_rollout}
\end{figure}

\begin{figure}[H]
  \centering
  \includegraphics[width=0.5\textwidth]{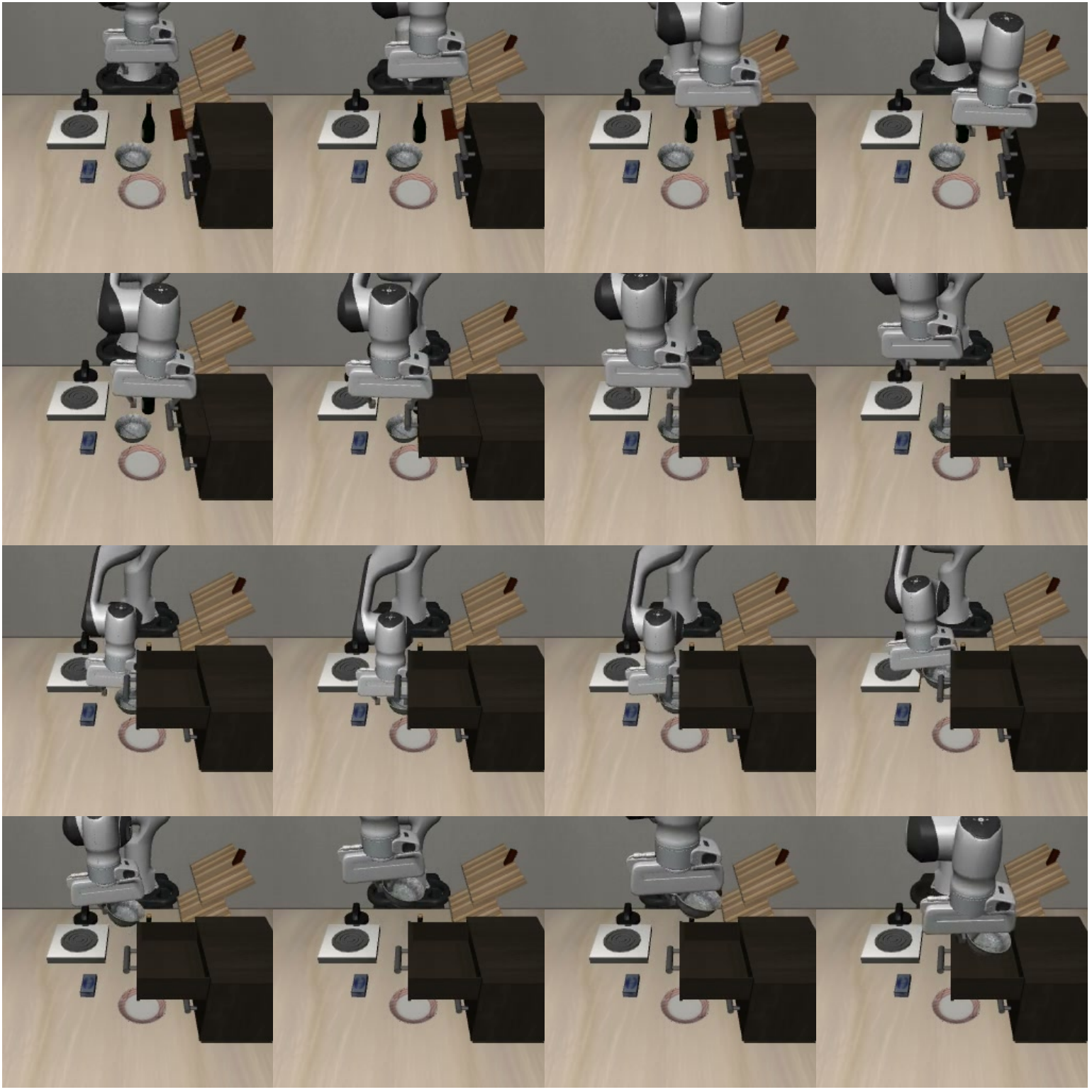}
  \caption{LIBERO
  \textit{open\_the\_top\_drawer\_and\_put\_the\_bowl\_inside}: a
  two-phase task (drawer opening, bowl transport and placement) under
  changing scene geometry.}
  \label{fig:libero_bowl_rollout}
\end{figure}

\begin{figure}[H]
  \centering
  \includegraphics[width=0.8\textwidth]{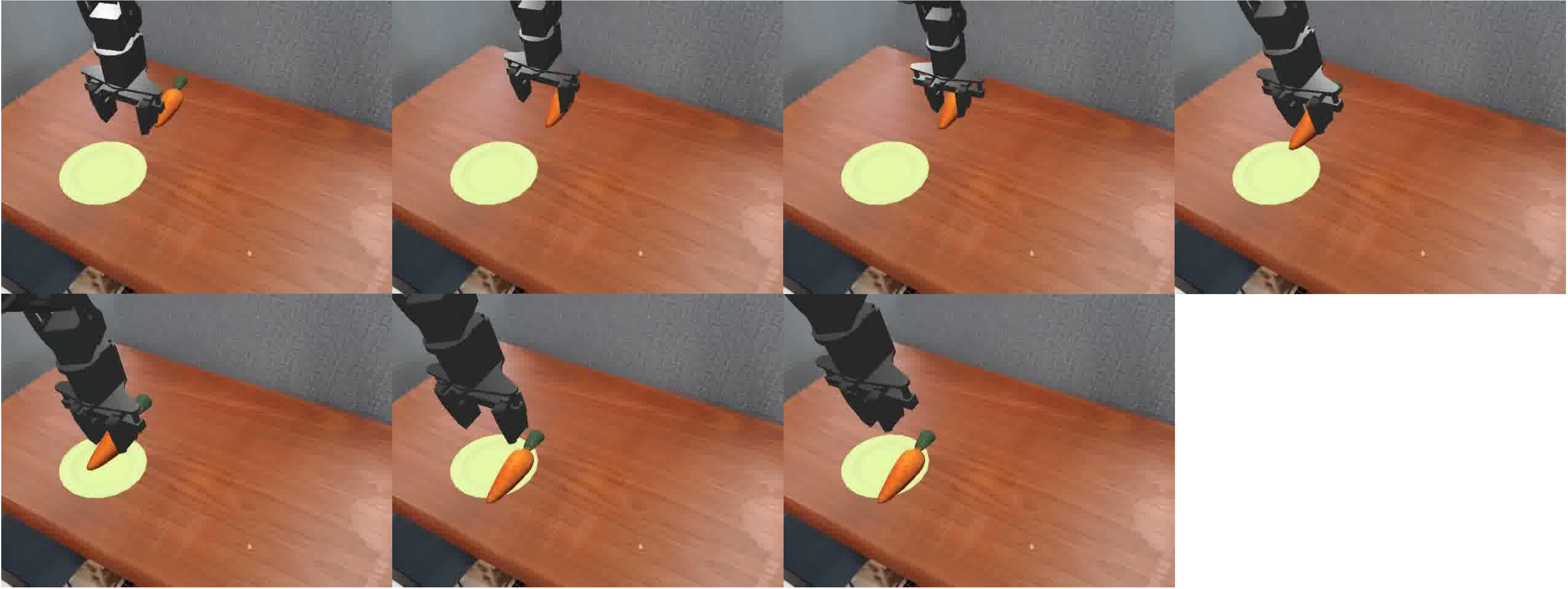}
  \caption{SimplerEnv-WidowX \textit{carrot}: stable grasp with
  consistent target alignment during transport.}
  \label{fig:widowx_carrot_rollout}
\end{figure}

\begin{figure}[H]
  \centering
  \includegraphics[width=0.8\textwidth]{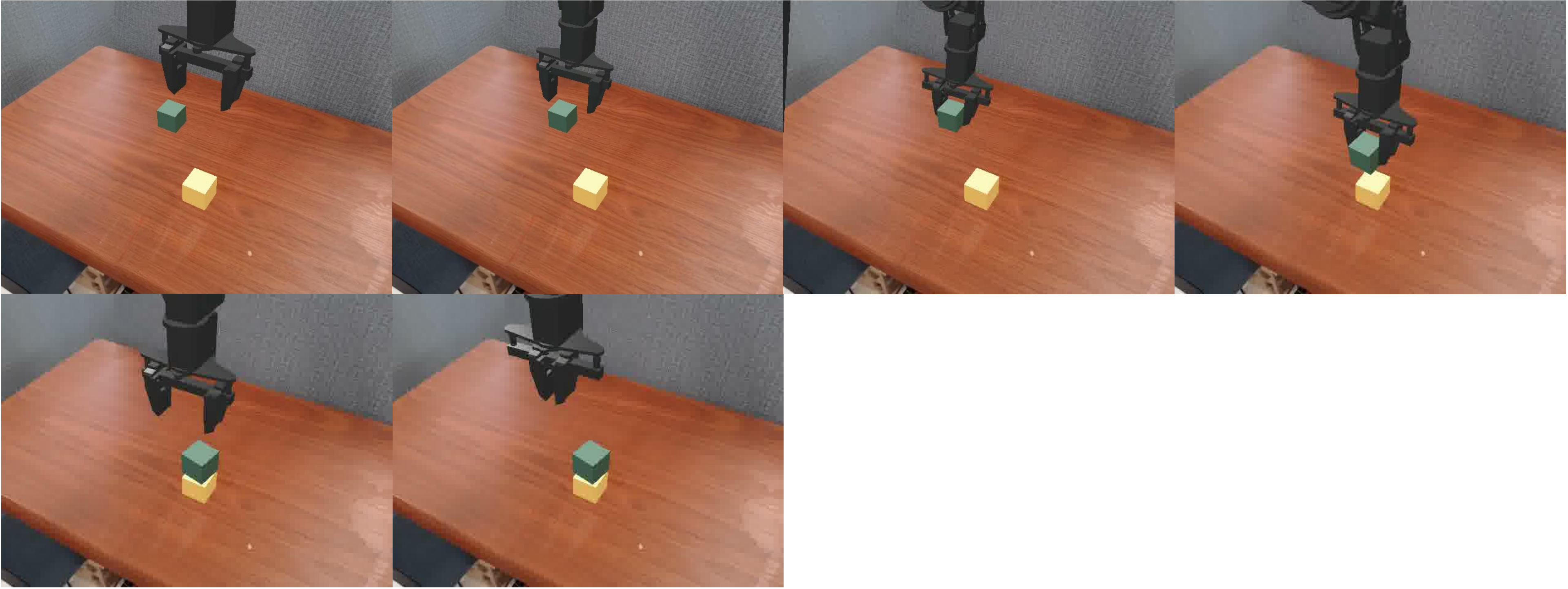}
  \caption{SimplerEnv-WidowX \textit{stack}: relative-pose alignment
  prior to final placement, illustrating spatial grounding in
  contact-sensitive stacking.}
  \label{fig:widowx_stack_rollout}
\end{figure}

\begin{figure}[H]
  \centering
  \includegraphics[width=0.8\textwidth]{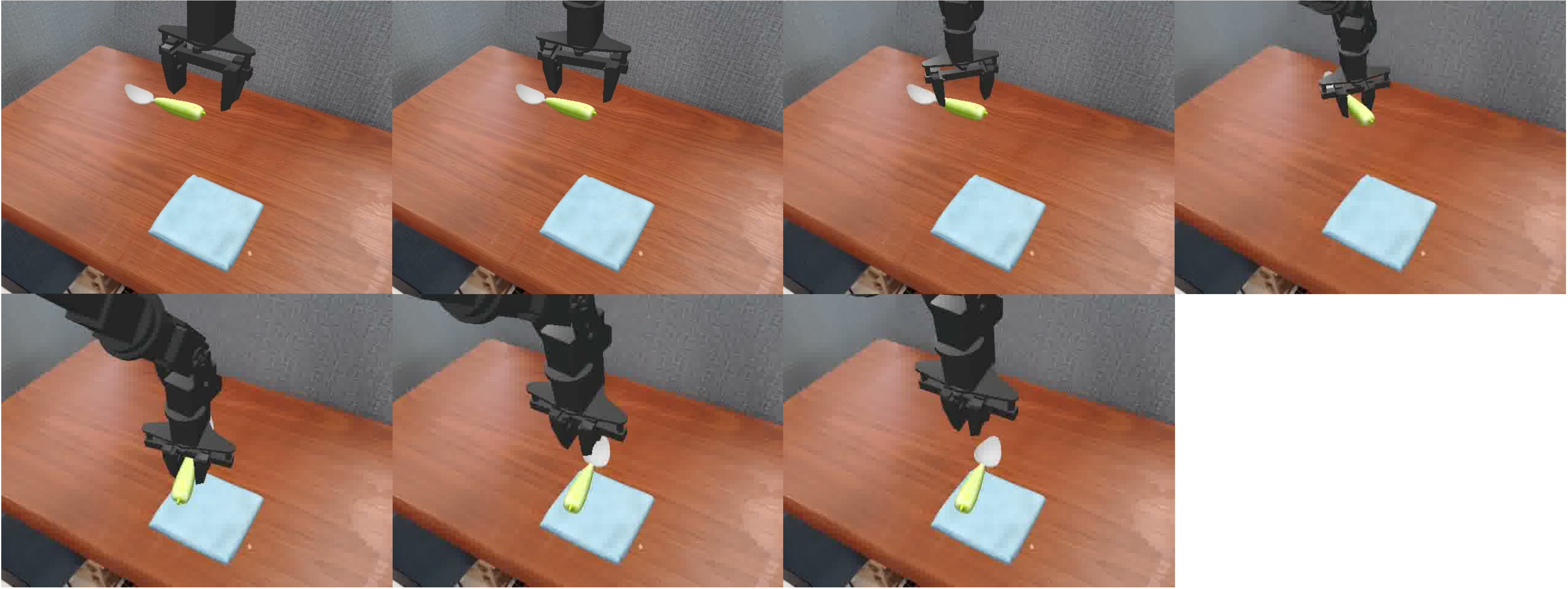}
  \caption{SimplerEnv-WidowX \textit{spoon}: stable grasp and
  placement of an elongated tool-shaped object under shape-dependent
  pose constraints.}
  \label{fig:widowx_spoon_rollout}
\end{figure}

Figs.~\ref{fig:libero_stove_rollout}--\ref{fig:widowx_spoon_rollout}
present rollout visualizations on LIBERO~\citep{liu2023libero} and
SimplerEnv-WidowX~\citep{li2024evaluating}
covering articulated interaction, multi-stage compositional tasks,
and fine-grained placement.

\subsection{Additional Real-World Qualitative Results}
\label{app:more_visualization}

\begin{figure}[H]
  \centering
  \includegraphics[width=0.6\linewidth]{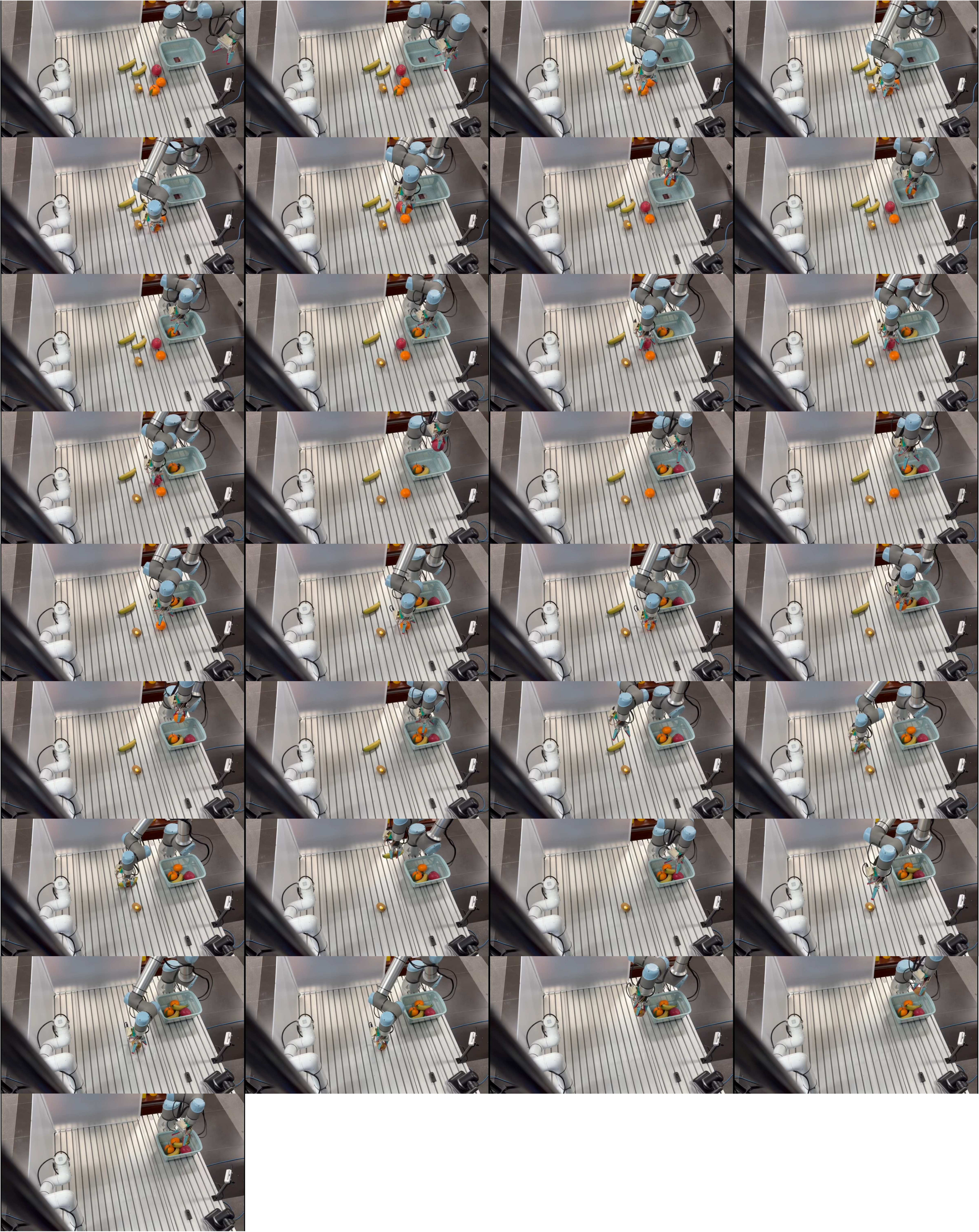}
  \caption{Representative \textbf{Fruit Collecting} rollout showing repeated grasp--transport--release cycles as the basket fills.}
  \label{fig:fruit_demo}
\end{figure}

\begin{figure}[H]
  \centering
  \includegraphics[width=0.6\linewidth]{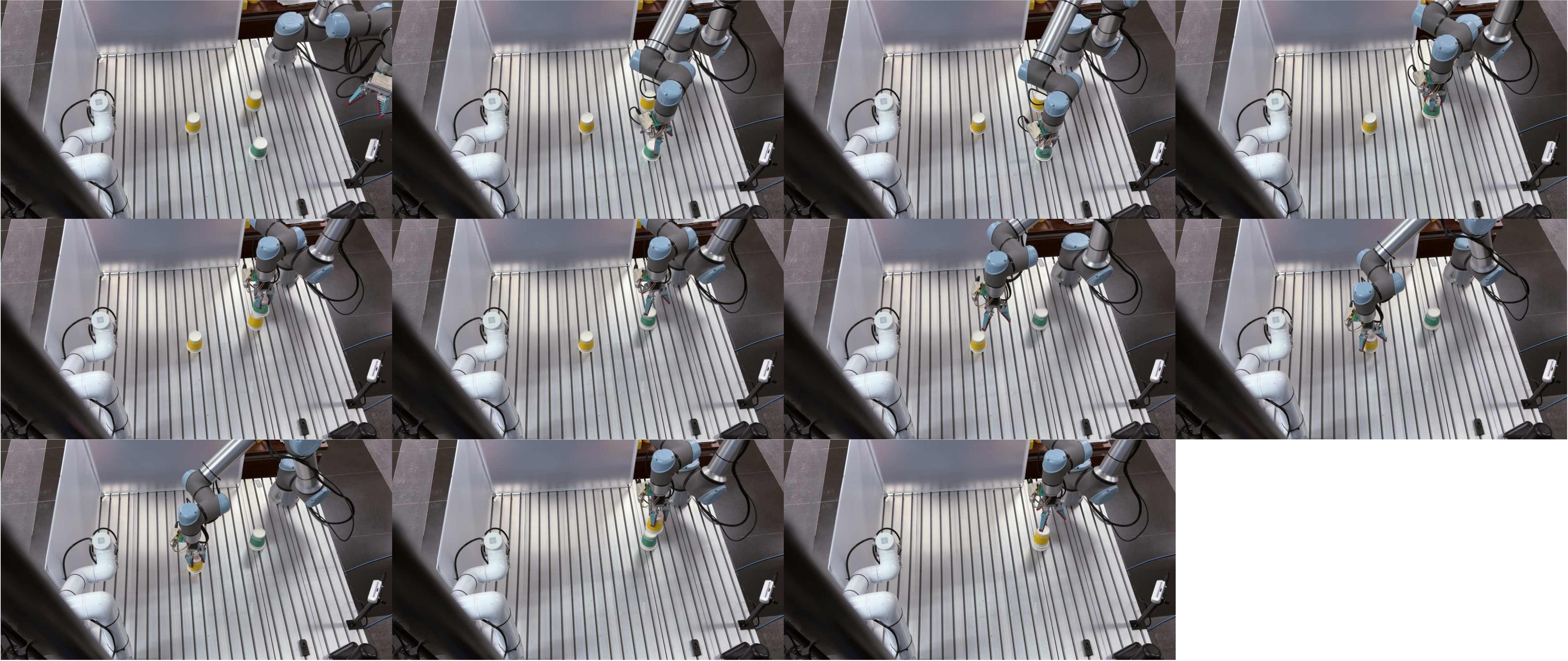}
  \caption{Representative \textbf{Paper Cup Stacking} rollout showing cup lifting, alignment, and placement.}
  \label{fig:cup_demo}
\end{figure}

\begin{figure}[H]
  \centering
  \includegraphics[width=0.6\linewidth]{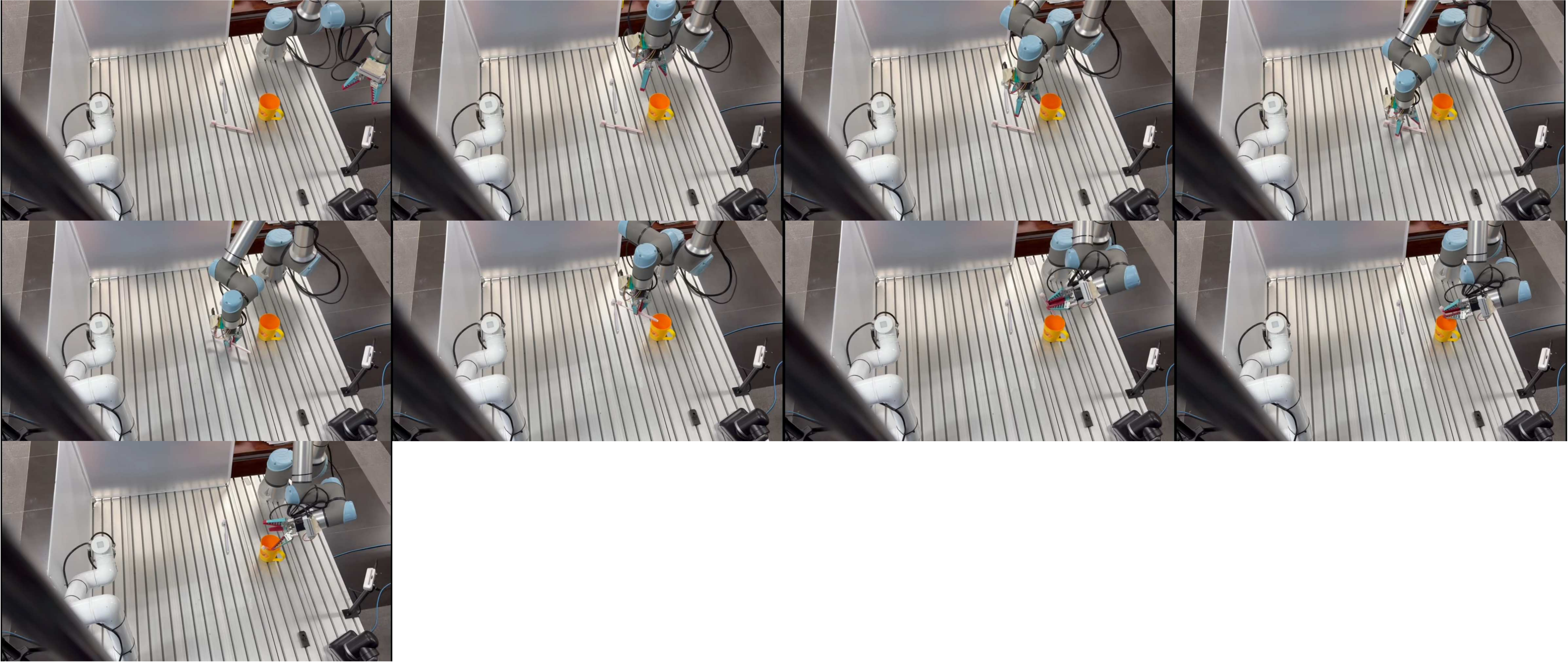}
  \caption{Representative \textbf{Toothbrush Insertion} rollout showing grasp, transport, pose adjustment, and insertion.}
  \label{fig:toothbrush_demo}
\end{figure}

\begin{figure}[H]
  \centering
  \includegraphics[width=0.6\linewidth]{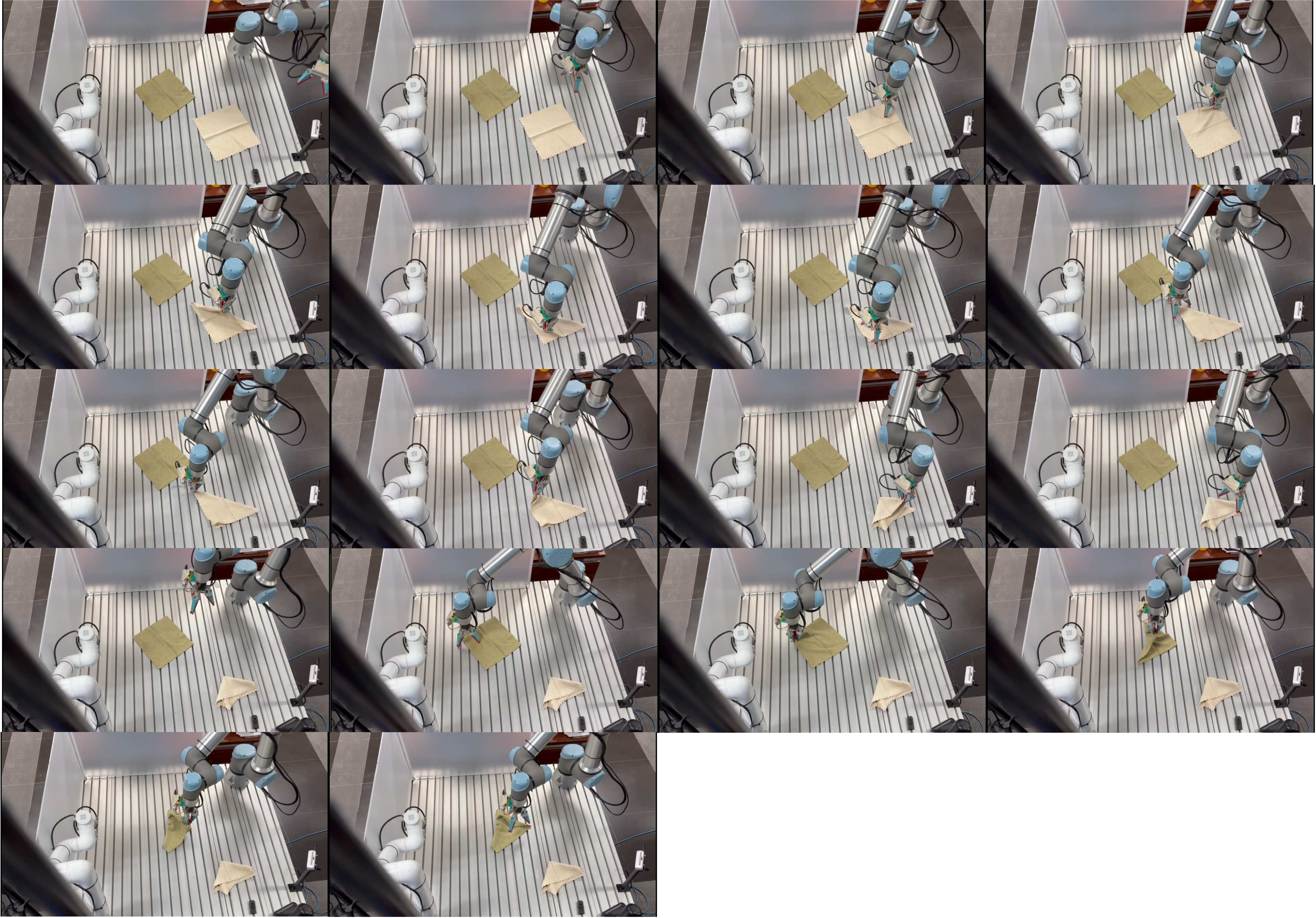}
  \caption{Representative \textbf{Cloth Folding} rollout showing pulling and folding as the cloth configuration changes.}
  \label{fig:cloth_demo}
\end{figure}

Figs.~\ref{fig:fruit_demo}--\ref{fig:cloth_demo} show third-person views
of representative executions of the four tasks.
Quantitative success and open-loop results are in
Appendix~\ref{app:realworld_stats}.

\paragraph{Failure examples.}
Figure~\ref{fig:realenv} includes Cloth Folding failures involving
misalignment and excessive force. Additional failure categories described
in the evaluation include premature release and incomplete folding.

\FloatBarrier
\section{Design Scope and Limitations}
\label{app:limitations}
\label{app:factorization_rationale}
SCULPT-VLA uses a fixed decomposition into task-progression,
scene-dynamics, and spatial-grounding factors, whose roles are established
by training-time teachers.

Physical evaluation covers four tabletop tasks under task-specific ID
and OOD conditions on one UR5e platform.

\end{document}